\documentclass{article}

\usepackage[preprint]{neurips_2025}

\usepackage[utf8]{inputenc} 
\usepackage[T1]{fontenc}    
\usepackage{hyperref}       
\usepackage{url}            
\usepackage{booktabs}       
\usepackage{amsfonts}       
\usepackage{nicefrac}       
\usepackage{microtype}      
\usepackage{xcolor}         
\usepackage{pifont}
\usepackage{graphicx}
\usepackage{float}
\usepackage{amsmath}
\usepackage{tabularx}
\usepackage{multirow}
\usepackage{scrextend}
\usepackage{algorithm}
\usepackage{algpseudocode}
\usepackage{multirow}
\usepackage{subcaption}
\usepackage{cellspace} 
\title{Bubbleformer: Forecasting Boiling with Transformers}
\author{%
Sheikh Md Shakeel Hassan$^{1}$\thanks{Corresponding authors: \texttt{\{sheikhh1,amowli\}@uci.edu}} \quad Xianwei Zou$^{1}$ \quad Akash Dhruv$^2$ \\ \quad \textbf{Viswanath Ganesan}$^3$ \textbf{Aparna Chandramowlishwaran}$^{1\ast}$ \\
$^1$University of California, Irvine \quad $^2$Argonne National Laboratory \quad $^3$UIUC}
\begin{document}

\maketitle

\begin{abstract}
    Modeling boiling---an inherently chaotic, multiphase process central to energy and thermal systems---remains a significant challenge for neural PDE surrogates. Existing models require future input (e.g., bubble positions) during inference because they fail to learn nucleation from past states, limiting their ability to autonomously forecast boiling dynamics. 
    They also fail to model flow boiling velocity fields, where sharp interface–momentum coupling demands long-range and directional inductive biases.
    
    We introduce \textbf{Bubbleformer}, a transformer-based spatiotemporal model that forecasts stable and long-range boiling dynamics including nucleation, interface evolution, and heat transfer without dependence on simulation data during inference.
    Bubbleformer integrates factorized axial attention, frequency-aware scaling, and conditions on thermophysical parameters to generalize across fluids, geometries, and operating conditions.
    To evaluate physical fidelity in chaotic systems, we propose interpretable physics-based metrics that evaluate heat flux consistency, interface geometry, and mass conservation.
    We also release \textbf{BubbleML 2.0}, a high-fidelity dataset that spans diverse working fluids (cryogens, refrigerants, dielectrics), boiling configurations (pool and flow boiling), flow regimes (bubbly, slug, annular), and boundary conditions. Bubbleformer sets new benchmark results in both prediction and forecasting of two-phase boiling flows. 
\end{abstract}

\section{Introduction}
\begin{figure}[h]
    \centering
    \includegraphics[width=1.0\linewidth]{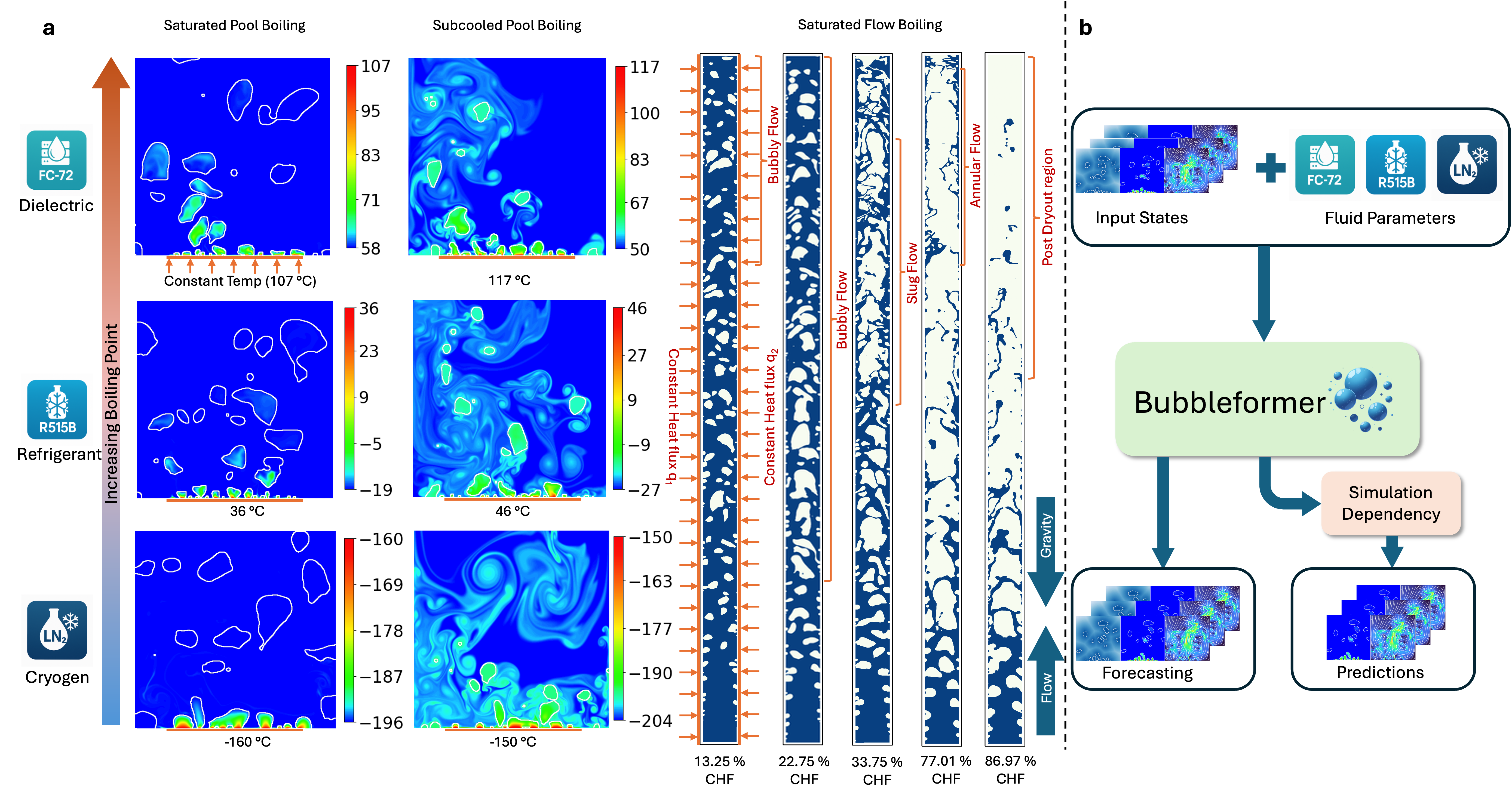}
    \caption{\textbf{(a) BubbleML 2.0 dataset}: Visualizes temperature and phase fields across pool and flow boiling configurations. Bubbles nucleate on heated surfaces, deform, coalesce, and transition between regimes, with dynamics strongly coupled to boundary conditions, working fluid, and geometry. \textbf{(b) Bubbleformer downstream tasks}: Autonomous \emph{forecasting} of full-field dynamics including nucleation and interface evolution, and \emph{prediction} conditioned on future bubble interfaces.}
    \label{fig:overview}
\end{figure}

Boiling is one of the most efficient modes of heat transfer due to the large latent heat of vaporization at liquid-vapor interfaces, making it an attractive solution for ultra-high heat flux applications such as nuclear reactors and next-generation computing infrastructure. 
Companies such as ZutaCore\footnote{\url{https://zutacore.com}} and LiquidStack\footnote{\url{https://www.liquidstack.com}} are pioneering two-phase and immersion cooling technologies for data centers supporting AI workloads. 
These industrial efforts reflect a broader trend toward harnessing phase-change phenomena for thermal control in compact, high-density environments. 
Boiling also holds promise for spacecraft thermal control, but without buoyancy to aid vapor removal, boiling in microgravity faces severe challenges, limiting its current viability.
Addressing these limitations will require novel boiling architectures optimized for low gravity, guided by modeling, design optimization, and experimental validation--an iterative process that is computationally intensive and costly. 
More fundamentally, accurately modeling two-phase pool and flow boiling remains one of the grand challenges in fluid dynamics. The underlying physics is inherently chaotic and multiscale: bubbles nucleate stochastically on heated surfaces, liquid-vapor interfaces continuously deform, coalesce, and break apart, and transitions between flow regimes (e.g., bubbly, slug, annular) occur unpredictably under strong thermal-hydrodynamic coupling. As illustrated in Figure~\ref{fig:overview}, boiling systems can span a vast range of spatial and temporal scales and are highly sensitive to boundary conditions, fluid properties, and geometry.

Recent advances in high-fidelity multiphysics solvers such as Flash-X \cite{DUBEY2022,DHRUV2024113122} have enabled simulations of boiling by solving incompressible Navier-Stokes equations in two phases with interface tracking and nucleation models. Although physically accurate, these simulations are computationally expensive, often requiring days on petascale supercomputers to simulate seconds of physical time \cite{DHRUV2019103099,DHRUV2021121826}. This has motivated the development of machine learning (ML) surrogates that learn to approximate spatiotemporal evolution of boiling directly from data. These models promise orders-of-magnitude acceleration, enabling new capabilities in real-time forecasting, parametric studies, and design-space exploration. However, current ML models for boiling \cite{hassan2023bubbleml, khodakarami2025mitigatingspectralbiasneural} exhibit three main limitations: (1) They require future bubble positions as input to predict velocity and temperature fields, making them unsuitable for forecasting, (2) They fail to learn bubble nucleation, a stochastic discontinuous process  central to long rollouts, and (3) They fail to predict velocity fields in flow boiling, even when provided with future bubble positions.

We introduce \textbf{Bubbleformer}, a transformer-based spatiotemporal model that forecasts full-field boiling dynamics that includes temperature, velocity, and signed distance fields representing interfaces, setting a new benchmark for ML-based boiling physics. Bubbleformer makes the following core contributions: 

\begin{itemize}
    \item \textbf{Beyond prediction to forecasting.} By operating directly on full 5D spatiotemporal tensors and preserving temporal dependencies, Bubbleformer learns nucleation, key to forecasting and predicting long-range dynamics. Unlike prior models that compress time or require future bubble positions, our approach infers them end-to-end.
    \item \textbf{Generalizing across fluids and flow regimes.} We introduce \textbf{BubbleML 2.0}, the most comprehensive boiling dataset to date, comprising over 160 high-fidelity simulations across diverse fluids (e.g., cryogenics, refrigerants, dielectrics), boiling configurations (pool and flow), heater geometries (single- or double-sided heating), and flow regimes (bubbly, slug, and annular until dryout).  
    Bubbleformer is conditioned on thermophysical parameters, allowing a single model to generalize across these axes.
    \item \textbf{Physics-based evaluation.} We introduce new interpretable metrics to assess physical fidelity beyond pixel-wise error. These include heat flux divergence, Eikonal equation loss for signed distance functions, and conservation of vapor mass. Together, these metrics provide a more rigorous evaluation of physical correctness in chaotic boiling systems.
\end{itemize}

To our knowledge, \emph{Bubbleformer is the first model to demonstrate autonomous, physically plausible forecasting of boiling dynamics}. It sets new state-of-the-art benchmarks on both prediction and forecasting tasks in BubbleML 2.0, representing a significant step toward practical, generalizable ML surrogates for multiphase thermal transport.
\section{Problem Statement}
Boiling involves the phase change of a liquid into vapor at a heated surface, forming bubbles that enhance turbulence and heat transfer. This phenomenon is highly chaotic: bubbles nucleate unpredictably on the heated surface, then grow, merge, and eventually detach, all while interacting with the surrounding liquid. 
The boiling system is governed by the incompressible Navier-Stokes equations (for momentum conservation) coupled with an energy transport equation, solved in both the liquid and vapor phases. 
These equations describe the evolution of the fluid’s velocity $\vec{u}$ and pressure fields $P$ along with the temperature field $T$ that captures heat distribution.
The \emph{liquid-gas interface} is represented by a level-set function $\phi$ (signed distance field) that tracks the moving phase boundary. The interface $\Gamma$ is defined by $\phi = 0$, with $\phi > 0$ in the vapor region and $\phi < 0$ in the liquid.
The governing equations are non-dimensionalized using characteristic scales defined in the liquid phase (e.g., capillary length for length scale and the terminal velocity for velocity scale) and key dimensionless numbers such as Reynolds, Prandtl, Froude, Peclet, and Stefan. To ensure consistency across both phases, vapor properties (e.g., density, viscosity, thermal conductivity, and specific heat) are specified relative to the liquid's properties.

Interfacial physics is modeled by enforcing the conservation of mass and energy at $\Gamma$, accounting for surface tension and latent heat. Jump conditions for velocity, pressure, and temperature across the interface are implemented using the Ghost Fluid Method (GFM), a numerical scheme that handles sharp discontinuities at the boundary without smearing them. 
The level-set interface $\phi$ is updated via an advection (convection) equation, which moves the interface with the local fluid velocity. Evaporation and condensation are governed by differences in interfacial heat flux between the two phases, coupling flow dynamics with thermal effects. The two-phase numerical simulation framework is implemented in Flash-X \cite{DUBEY2022} and follows the formulation in \cite{DHRUV2024113122}. For completeness, the governing equations and numerical modeling assumptions are provided in Appendix \ref{app:numerical}.
\section{Failure Modes of Neural Solvers for Boiling}

Recent works \cite{hassan2023bubbleml, khodakarami2025mitigatingspectralbiasneural} have explored neural surrogates trained on Flash-X simulations \cite{DUBEY2022,DHRUV2024113122,DHRUV2019103099} to model boiling dynamics. 
These models aim to predict future velocity, temperature, and interface evolution given past physical states. 
While promising, current architectures exhibit persistent failure modes that limit their ability to forecast real-world systems.

\subsection{Simulation Dependency and Failure to Learn Nucleation} 
\label{subsec:failmode-nucleate}

Current boiling models are trained to predict velocity and temperature fields given both past physical states and \emph{future} bubble positions. 
A task in BubbleML \cite{hassan2023bubbleml} is to learn an operator: $\mathcal{G}(\phi_{prev}, \vec{u}_{prev}, T_{prev}, \phi_{next}) = [\vec{u}_{next}, T_{next}]$. 
Although this task is tractable during supervised training, it introduces a serious limitation during inference: the model requires access to $\phi_{next}$, which are not available without running the underlying simulation. 
This precludes autonomous rollouts and the model is fundamentally unsuitable for forecasting.

\begin{figure}[h]
    \centering
    \includegraphics[width=1.0\linewidth]{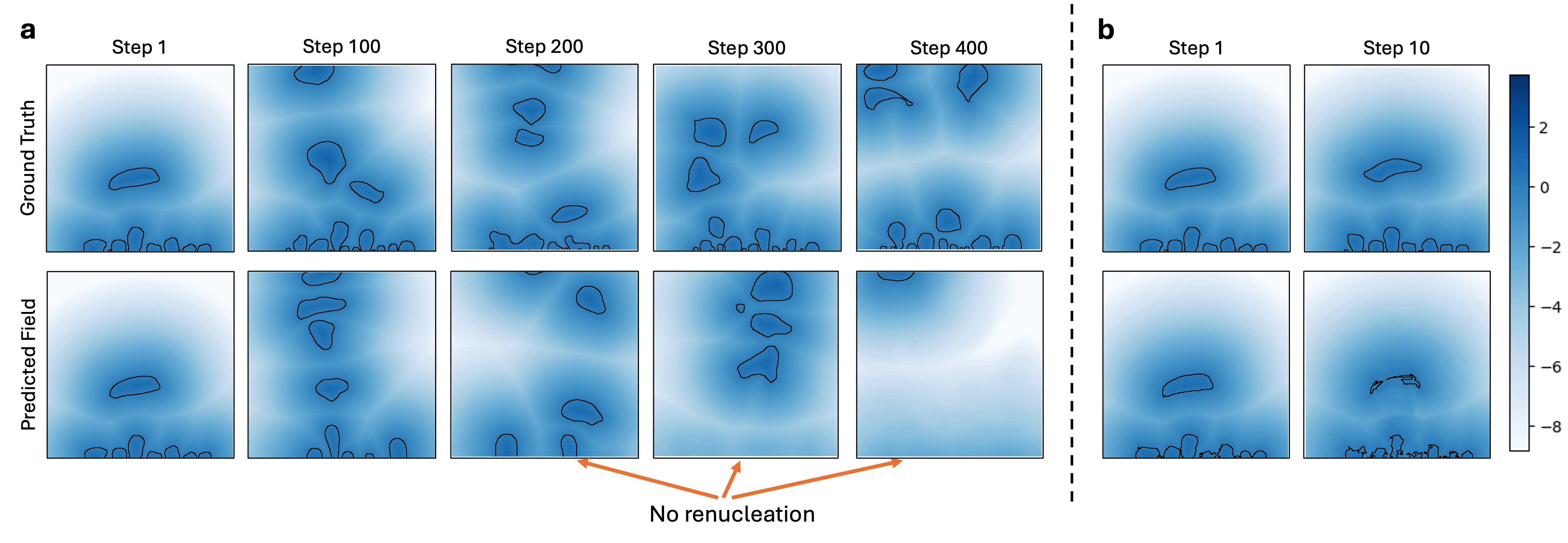}
    \caption{\textbf{Failure modes of current ML models in forecasting.} Left: UNet models maintain smooth evolution during autoregressive rollout but fail to nucleate. Right: FNO exhibits early instability.}
    \label{fig:failure-modes}
\end{figure}

To eliminate this dependency, one can attempt to jointly learn bubble evolution: $\mathcal{G}(\phi_{prev}, \vec{u}_{prev}, T_{prev}) = [\phi_{next}, \vec{u}_{next}, T_{next}]$, requiring the model to predict future nucleation events purely from historical data. 
However, this task proves challenging for current architectures.
Bubble nucleation is a discontinuous stochastic phenomena governed by microscale surface physics, contact angles, and thermal boundary layers \cite{GILMAN201735,Yazdani2016,DHRUV2019103099}. 
In Flash-X, nucleation is modeled algorithmically: new bubbles reappear on the heater surface at nucleation sites with seed radii after specified wait time (time before a new bubble forms after the prior bubble departs).
When these conditions are satisfied, re-nucleation is achieved through the union of the new signed distance field (due to the newly formed bubble) with the evolving phase field, introducing sharp discontinuities and topological changes. 

These reinitializations are non-differentiable and violate assumptions of spatial smoothness. 
As a result, learning this behavior directly from data proves difficult. 
We train UNet and Fourier Neural Operator (FNO) models from BubbleML \cite{hassan2023bubbleml} to learn this task. 
UNet-based models perform well in single-step prediction but fail in autoregressive rollouts, as they do not learn to nucleate new bubbles as shown in Figure \ref{fig:failure-modes}.
FNO models degrade more rapidly, often diverging after a single step. FNO relies on global spectral filters and smooth continuous mappings between input and output Banach spaces\cite{kovachki2021neural}. 
The sharp discontinuities associated with nucleation add "shocks" in the underlying function space, making it difficult for the neural operator models leading to quick divergence during rollout. This is consistent with theoretical analysis showing that neural operators are sensitive to abrupt and high-frequency discontinuous topological changes\cite{chauhan2025neuraloperatorsstrugglelearn}\cite{lanthaler2023nonlinear}.

\subsection{Failure in Flow Boiling Velocity Prediction}
\label{subsec:failmode-fbvel}

Flow boiling introduces additional modeling challenges. 
Unlike pool boiling, which is largely buoyancy-driven and symmetric, flow boiling features directional velocity, thin films, and shear-induced instabilities. 
Accurate modeling of such flows requires resolving the tight coupling between evolving interface dynamics and momentum transport (i.e., velocity fields).
We observe that UNet and FNO models fail to predict velocity in flow boiling datasets. 
This failure occurs despite training on fine-resolution data and using future interface fields as supervision.
We identify two primary contributing factors:

\textbf{Lack of directional inductive bias.} Flow boiling geometries are typically long, narrow rectangular domains (see Figure \ref{fig:overview}), with a higher resolution in the flow direction than in the cross-stream direction.
When FNO's low-pass filters are applied isotropically in Fourier space, they disproportionately suppress high-frequency modes along the flow-direction, resulting in directional aliasing.
Similarly, UNet models apply symmetric convolutions across both axes and lack the architectural bias to prioritize dominant features, limiting their ability to resolve streamwise velocity gradients and anisotropic flow patterns.

\textbf{Insufficient spatiotemporal integration.} Unlike pool boiling, flow boiling introduces additional spatiotemporal gradients from the bulk liquid flow. 
Localized changes in interface topology (e.g., film rupture, bubble coalescence) induce sharp and nonlocal responses in the velocity field \cite{ganesan2018development}. 
Predicting this behavior requires long-range spatial context and multiscale temporal integration. 
UNet models, while effective for local pixelwise regression, lack sufficient temporal memory and spatial receptive field to capture these dynamics. 
FNOs, on the other hand, suffer from spectral oversmoothing, which makes it difficult to isolate sharp interface-driven effects.

These failures are not isolated. Across all flow boiling datasets in BubbleML, we observe that FNO and UNet models fail to converge when learning to predict velocity. This highlights the need for architectures that go beyond spectral operators and spatial attention to incorporate hierarchical, directional, and temporally-aware representations.
\section{Bubbleformer}

Bubbleformer is a spatiotemporal transformer designed to forecast boiling dynamics across fluids, boiling configurations, geometries, and flow regimes.
In contrast to prior surrogates that fail to nucleate (Section \ref{subsec:failmode-nucleate}) or generalize to flow boiling (Section \ref{subsec:failmode-fbvel}), Bubbleformer integrates structural innovations that enable long rollout, parameter generalization, and high-frequency prediction. 

\subsection{Model Architecture}

\begin{figure}[h]
    \centering
    \includegraphics[width=1.0\linewidth]{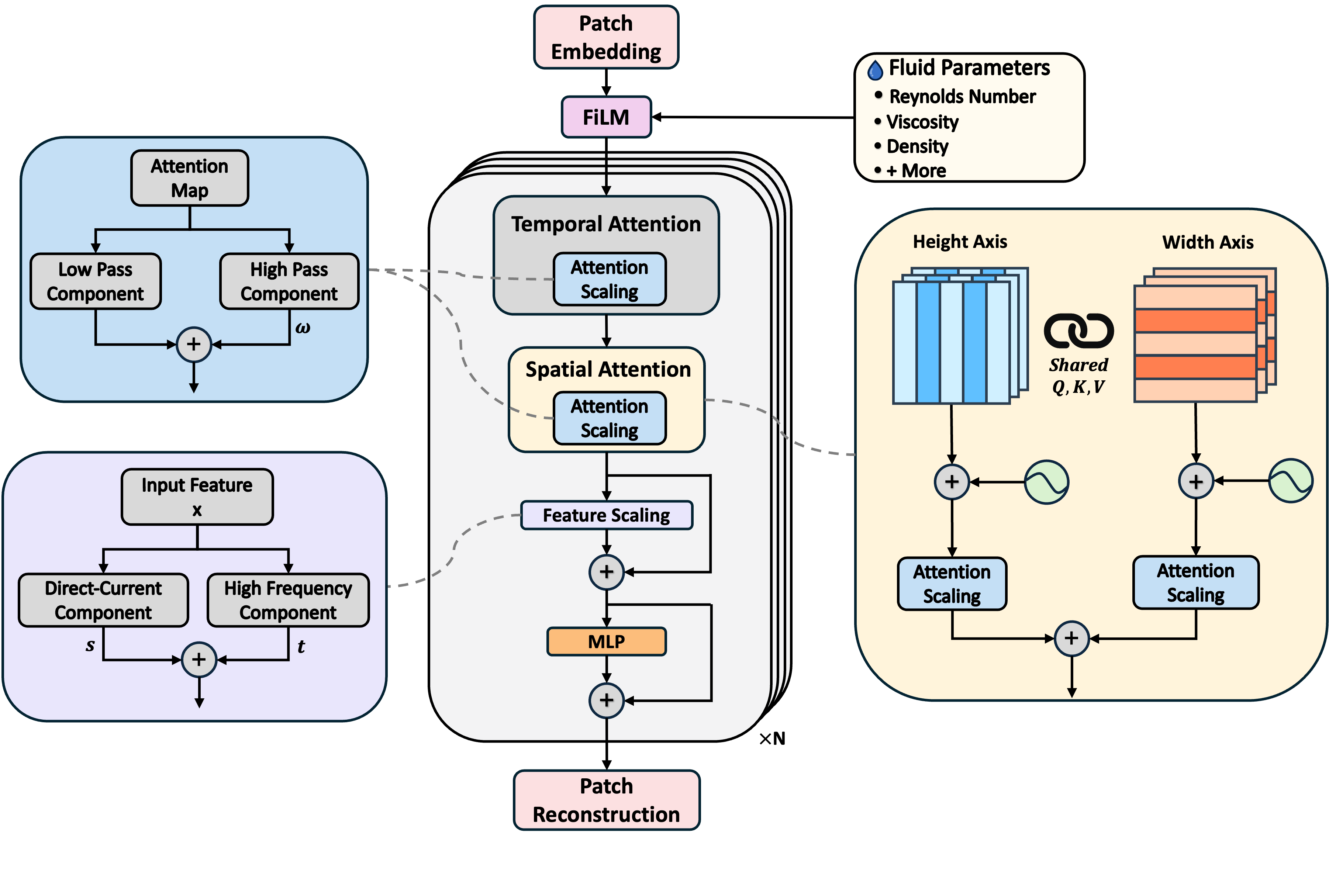}
    \caption{\textbf{Bubbleformer architecture.} Input fields are encoded into multiscale spatiotemporal patches, conditioned with fluid-specific parameters, transformed by frequency-scaled temporal and spatial axial attention, and decoded into future velocity, temperature, and interface fields.}
    \label{fig:bubbleformer}
\end{figure}

Bubbleformer architecture illustrated in Figure \ref{fig:bubbleformer} consists of four components: hierarchical patch embedding, physics-based parameter conditioning, factorized space-time axial attention, and frequency-aware feature modulation.

\textbf{Hierarchical Patch Embedding.} 
Bubbleformer applies a hierarchical MLP (hMLP) stem inspired by recent hybrid vision transformers \cite{touvron2022thingsknowvisiontransformers} to embed input physical fields as spatiotemporal patches.
Each time slice is processed using a series of non-overlapping 2$\times$2 convolutional layers with stride 2, progressively reducing spatial resolution and increasing representational depth. 
This design differs from prior hMLP implementations that fix resolution to 16$\times$16 patches \cite{touvron2022thingsknowvisiontransformers, mccabe2024multiple}.
Instead, we stack repeated 2$\times$2 kernels at each stage to support flexible generalization across multiple resolutions.
Compared to flat patch tokenization in ViT \cite{dosovitskiy2020vit}, this embedding builds a multiscale feature hierarchy with a stronger inductive bias for boiling dynamics, especially in domains with high aspect-ratio and varying discretizations.
    
\textbf{FiLM-based Parameter Conditioning.} 
To generalize across fluids (such as cryogens, refrigerants, and dielectrics), Bubbleformer conditions embedded feature representations on physical parameters.
A feature-wise linear modulation (FiLM) layer \cite{perez2017filmvisualreasoninggeneral} applies a learned affine transformation on each channel of the patch embeddings using coefficients derived from a 9D fluid descriptor: Reynolds number, Prandtl number, Stefan number, viscosity, density, thermal conductivity, specific heat capacity, heater temperature and nucleation wait time. 
A small MLP maps this descriptor to channel-wise scaling and bias terms ($\gamma_c$, $\beta_c$), which are applied as: $f’_c = \gamma_c f_c + \beta_{c}$ for each channel $c$.
This conditioning allows the model to adapt its internal representations to fluid-specific thermophysical properties, essential for generalization across working fluids.
Without such conditioning, the model cannot distinguish between fluids with drastically different boiling characteristics (see Figure \ref{fig:overview}) resulting in mispredicted nucleation timing, incorrect interface velocities, and failure to preserve heat flux scaling during inference.
    
\textbf{Factorized Space-Time Axial Attention.}
To model long-range spatiotemporal dynamics in boiling flows, Bubbleformer combines \emph{factorized space-time attention} \cite{gberta_2021_ICML, arnab2021vivit} with \emph{axial attention} \cite{ho2019axialattentionmultidimensionaltransformers, huang2019ccnet, wang2020axial}.
In factorized space-time attention, temporal attention is applied first independently at each spatial location followed by spatial attention. 
This preserves the spatial locality during temporal modeling and has proven effective in video transformers \cite{gberta_2021_ICML, arnab2021vivit}.
At each Bubbleformer block, temporal attention operates across input timesteps to learn dynamics such as bubble growth, departure, and renucleation. 
The resulting encoding from temporal attention is then passed to spatial attention.
Instead of 2D self-attention, axial attention further decomposes the temporally updated features into two 1D attentions along the height and width axes.
This decomposition reduces the overall computational complexity from $\mathcal{O}(HWT)^2$ in joint space-time attention to $\mathcal{O}(H^2 + W^2 + T^2)$, while still maintaining a global receptive field.
This design, successfully applied to PDE forecasting \cite{mccabe2024multiple, zhang2024mateymultiscaleadaptivefoundation} addresses limitations in prior architectures. Direction-aware attention captures anisotropic structure--important for flow boiling--where strong gradients develop in the flow direction. Unlike FNO, which mixes spatial modes globally, axial attention preserves local structure and mitigates spectral oversmoothing.
Each axis learns shared query, key, and value projections, and we apply T5 relative position embeddings \cite{raffel2023exploringlimitstransferlearning} along both temporal and spatial attention blocks. 

\textbf{Attention and Frequency Scaling.} 
To prevent loss of detail in deeper layers, Bubbleformer incorporates frequency-aware attention and feature scaling \cite{wang2022antioversmooth}.
Deep transformers often suffer from attention collapse, where repeated application of softmax attention suppresses high-frequency signals, effectively acting as a low-pass filter. 
This degradation is particularly detrimental in boiling with high-frequency features such as sharp interfaces and condensation vortices. Attention scaling helps mitigate this by modifying attention to act more like an all-pass filter, by separately scaling the low- and high-frequency components of the attention scores. 
In addition, we also add a feature scaling layer \cite{wang2022antioversmooth} to each spatiotemporal block that explicitly separates the output feature map into low- and high-frequency components, reweights them with separate learnable parameters before recombining them. 
This acts as an adaptive sharpening filter that helps preserve fine-grained structures essential for modeling phase interfaces and turbulence. Similar high-frequency feature scaling has been applied successfully to improve temperature field prediction in BubbleML datasets using ResUNet and diffusion models \cite{khodakarami2025mitigatingspectralbiasneural}.

\textbf{Patch Reconstruction.}
The patch reconstruction mirrors hierarchical patch embedding in reverse. 
The spatiotemporal output embeddings are progressively upsampled through $k$ transposed convolution layers to recover the original spatial resolution.
This reconstruction produces future predictions of all physical fields--temperature, velocity, and signed distance--at each output timestep.

\subsection{Metrics}
We adopt and extend the metrics introduced in BubbleML \cite{hassan2023bubbleml} to evaluate both short-term predictive accuracy and long-term physical fidelity.
BubbleML includes field-based metrics, such as root mean squared error (RMSE), maximum squared error, relative L2 error, boundary RMSE (BRMSE), RMSE along bubble interfaces (IRMSE), and low/mid/high Fourier mode errors. 
These metrics provide a comprehensive view of spatial and frequency-domain accuracy, particularly useful for evaluating sharp gradients (e.g., temperature discontinuities across liquid-vapor boundaries) that may be masked by global averaging.
To evaluate Bubbleformer's long-horizon forecasting, we introduce three additional physically interpretable system-level metrics.

\textbf{Heat Flux Consistency.}
Boiling is inherently chaotic, and Flash-X simulations represent one possible trajectory of bubble dynamics under given boundary conditions among many. 
Small deviations in predicted bubble dynamics may lead to increasingly dissimilar fields, yet still preserve physically plausible behavior.
To assess system-level consistency, we measure the heat flux distribution across the heater surface over time.

In thermal science, \emph{heat flux} is a system-level indicator of boiling efficiency and its peak value--\emph{critical heat flux} (CHF)--marks the transition to boiling crisis due to the formation of a vapor barrier (see Figure \ref{fig:overview}). 
Accurate prediction of heat flux is essential for safe and efficient design of heat transfer systems \cite{liang2018pool, rassoulinejad2021deep, zhao2020prediction, sinha2021deep}.
We compute heat flux normal to the heater surface using Fourier's law \cite{incropera2007fundamentals}: 
\begin{equation}
q^j = k_\ell \left. \frac{\partial T^j}{\partial n} \right|_{\text{wall}}, \quad j \in \{t, \dots, t + k\} \nonumber
\end{equation}
where $k_l$ is the thermal conductivity of the liquid and $T$ is the temperature field. 
We accumulate heat flux over time, estimate its empirical distribution ($P_{\text{GT}}(q)$ and $P_{\text{ML}}(q)$) using kernel density estimation, and report mean, standard deviation, and Kullback-Leibler (KL) divergence \cite{Bishop:DeepLearning24} between simulation and model distributions:
\begin{equation}
D_{KL}(P_{\text{GT}}\,\|\,P_{\text{ML}}) = \int P_{\text{GT}}(q)\,\log\frac{P_{\text{GT}}(q)}{P_{\text{ML}}(q)}\,\mathrm{d}q \nonumber
\end{equation}

\textbf{Eikonal Loss.}
Phase interfaces (i.e., bubble positions) are represented via a signed distance field $\phi(x, y)$, which must satisfy the Eikonal equation: $|\nabla \phi| = 1$ throughout the domain. 
To assess geometric correctness of the predicted interfaces, we compute the pointwise Eikonal residual \cite{NEURIPS2023_2d6336c1} and report its average across all $N$ grid points:

\begin{equation}
    \mathcal{L}_{eik}(\phi) = \frac{1}{k}\sum_{j=t}^{t+k}\frac{1}{N}\sum_{i=1}^N\bigl||\nabla \phi^j_{\text{ML}}(x_i)|-1\bigr|
\end{equation}
A low Eikonal loss indicates that predicted interfaces preserve the level set property of the signed distance field and conform to physically valid bubble geometries. 
In practice, we find that losses below 0.1 threshold are sufficient to ensure stable interface evolution under autoregressive rollout.

\textbf{Mass Conservation.}
In boiling systems with fixed heater temperature, fluid properties, and surface geometry, the total vapor volume should remain approximately conserved, up to fluctuations from nucleation and condensation. 
We assess this, we compute the deviation in vapor volume between model predictions and ground truth simulations across the rollout window $[T, T+k]$.
We define the relative vapor volume error as:
\begin{equation}
\mathcal{L}_{vol}(\phi) = \frac{1}{k}\sum_{j=t}^{t+k}\frac{\bigl | \sum_{i=1}^N\mathbf{1}_{\{\phi_{\text{ML}}^j(x_i)>0\}} - \sum_{i=1}^N\mathbf{1}_{\{\phi_{\text{GT}}^j(x_i)>0\}} \bigr |}{\sum_{i=1}^N\mathbf{1}_{\{\phi_{\text{GT}}^j(x_i)>0\}}} \nonumber
\end{equation}

where $\mathbf{1}_{\{\phi(x)>0\}}$ is an indicator function for the vapor region, inferred from the signed distance field $\phi(x)$ at each timestep. 
A low $\mathcal{L}_{vol}(\phi)$ indicates that the model conserves global vapor volume in the domain.

\section{BubbleML 2.0 Dataset}

BubbleML 2.0 expands the original BubbleML dataset \cite{hassan2023bubbleml} with new fluids, boiling configurations, and flow regimes, enabling the study of generalization across thermophysical conditions and geometries. 
It adds \textbf{over 160 new high-resolution 2D simulations} spanning pool boiling and flow boiling, with diverse physics including saturated, subcooled, and single-bubble nucleation across three fluid types: FC-72 (dielectric), R-515B (refrigerant), and LN$_2$ (cryogen).
In addition to fluid diversity, BubbleML 2.0 introduces new constant heat flux boundary conditions with double-sided heaters to simulate different boiling regimes, including bubbly, slug, and annular until dryout.
Simulating these phenomena required advances in numerical methods in Flash-X. Appendix \ref{app:numerical} provides a detailed description of the simulations and validates against experimental data.

Table~\ref{tab:datset-summary} summarizes the dataset. 
All simulations are performed using Flash-X and stored in HDF5 format. 
Spatial and temporal resolution vary across fluids based on differences in characteristic scales, and adaptive mesh refinement (AMR) is used where needed for computational efficiency. Simulations performed on AMR grids are interpolated to regular grids with the same discretization as the other datasets.
We first apply linear interpolation, followed by nearest-neighbor interpolation to resolve boundary NaN values.
Additional details, including boundary conditions and data access instructions, are provided in Appendix~\ref{app:dataset-details}.

\begin{table}[h] \label{tab:dataset_summary}
  \caption {\textbf{Summary of BubbleML 2.0 datasets and their parameters.} $\Delta t$ is the temporal resolution in non-dimensional time which depends on the characteristic length and velocity for each fluid, as calculated in Table \ref{tab:simulation_parameters}. PB: pool boiling. FB: flow boiling.}
  \label{tab:datset-summary}
  \centering
  \newcommand{\results}[7]{ #1 & #2 & #4 & #3 & #6 & #5 & #7\\}

  \begin{tabular}{lllllll}
    \toprule
     Type - Physics - Fluid & Sims & Domain  & \multicolumn{2}{c}{Resolution} & Time  & Size \\
     & & ($mm^d$) & Spatial & $\Delta t$ & steps & (GB) \\
    \midrule
     \results{PB - Single Bubble - FC72}{11}{$192 \times 288$}{$4.38 \times 6.57$}{2000}{0.2}{50}
     \results{PB - Single Bubble - R515B}{11}{$192 \times 288$}{$6.48 \times 9.72$}{2000}{0.2}{50}
     \results{PB - Saturated - FC72}{20}{$512 \times 512$}{$11.68 \times 11.68$}{2000}{0.1}{320}
     \results{PB - Saturated - R515B}{20}{$512 \times 512$}{$17.28 \times 17.28$}{2000}{0.1}{320}
     \results{PB - Saturated - LN2}{20}{$512 \times 512$}{$16.96 \times 16.96$}{2000}{0.1}{320}
     \results{PB - Subcooled - FC72}{20}{$512 \times 512$}{$11.68 \times 11.68$}{2000}{0.1}{320}
     \results{PB - Subcooled - R515B}{20}{$512 \times 512$}{$17.28 \times 17.28$}{2000}{0.1}{320}
     \results{PB - Subcooled - LN2}{20}{$512 \times 512$}{$16.96 \times 16.96$}{2000}{0.1}{320}
     \results{FB - Inlet Velocity - FC72}{15}{$1344 \times 160$}{$30.66 \times 3.65$}{2000}{0.1}{320}
     \results{FB - Constant Heat Flux - FC72}{6}{$5152 \times 224$}{$117.5 \times 5.1$}{1000}{0.1}{420}
    \bottomrule
  \end{tabular}
\end{table}

BubbleML 2.0 follows FAIR data principles \cite{wilkinson2016fair} and includes metadata and physical parameters for every simulation as outlined in Appendix \ref{appendix:bml-fair}. The dataset enables benchmarking of generalization across fluids, boundary conditions, flow regimes, and supports training long-horizon forecasting models such as Bubbleformer.

\section{Results and Discussion}

\subsection{Training}
We train Bubbleformer to forecast future physical fields in boiling systems, including bubble positions (signed distance field $\phi$), temperature field ($T$), and velocity vector field ($\vec{u}$). 
Given $k$ past frames $[\phi, T, \vec{u}]_{t-k\,:\,t-1}$, the model predicts the next $k$ frames $[\phi, T, \vec{u}]_{t \,:\, t+k-1}$ in a bundled fashion \cite{brandstetter2022message}. 
Models are trained in a supervised manner by minimizing the sum of relative L2 norms across the predicted physical fields, $[\phi^*, \vec{u}^*, T^*]_{t \, : \, t+k-1}$. The single-step loss is given by:

\begin{equation}
\mathcal{L} = \frac{1}{k}\sum_{i=t}^{t+k-1}  \left( \frac{||\phi_i - \phi_i^*||_2}{||\phi_i||_2} + \frac{||T_i - T_i^*||_2}{||\phi_i||_2} + \frac{||\vec{u}_i - \vec{u}_i^*||_2}{||\vec{u}_i||_2}\right) \nonumber
\end{equation}
    
We train two Bubbleformer models, \emph{small} (29M parameters) and \emph{large} (115M parameters).  
using a prediction window of $k=5$ steps and train for 250 epochs using the Lion optimizer \cite{chen2023symbolicdiscoveryoptimizationalgorithms} and a warm-up cosine scheduler. 
Additional architectural and training details, including hyperparameter settings, are provided in Appendix \ref{app:forecast}.

\subsection{Re-nucleation and Forecasting}
In autoregressive forecasting, a basic requirement for all models is the nucleation of new bubbles on the heater surface when old ones depart. 
While prior models fail to renucleate, Bubbleformer successfully learns this behavior, capturing the temporal dynamics required for periodic bubble formation and maintaining stable rollouts over extended horizons. 

\begin{figure}[h]
\centering
\includegraphics[width=1.0\linewidth]{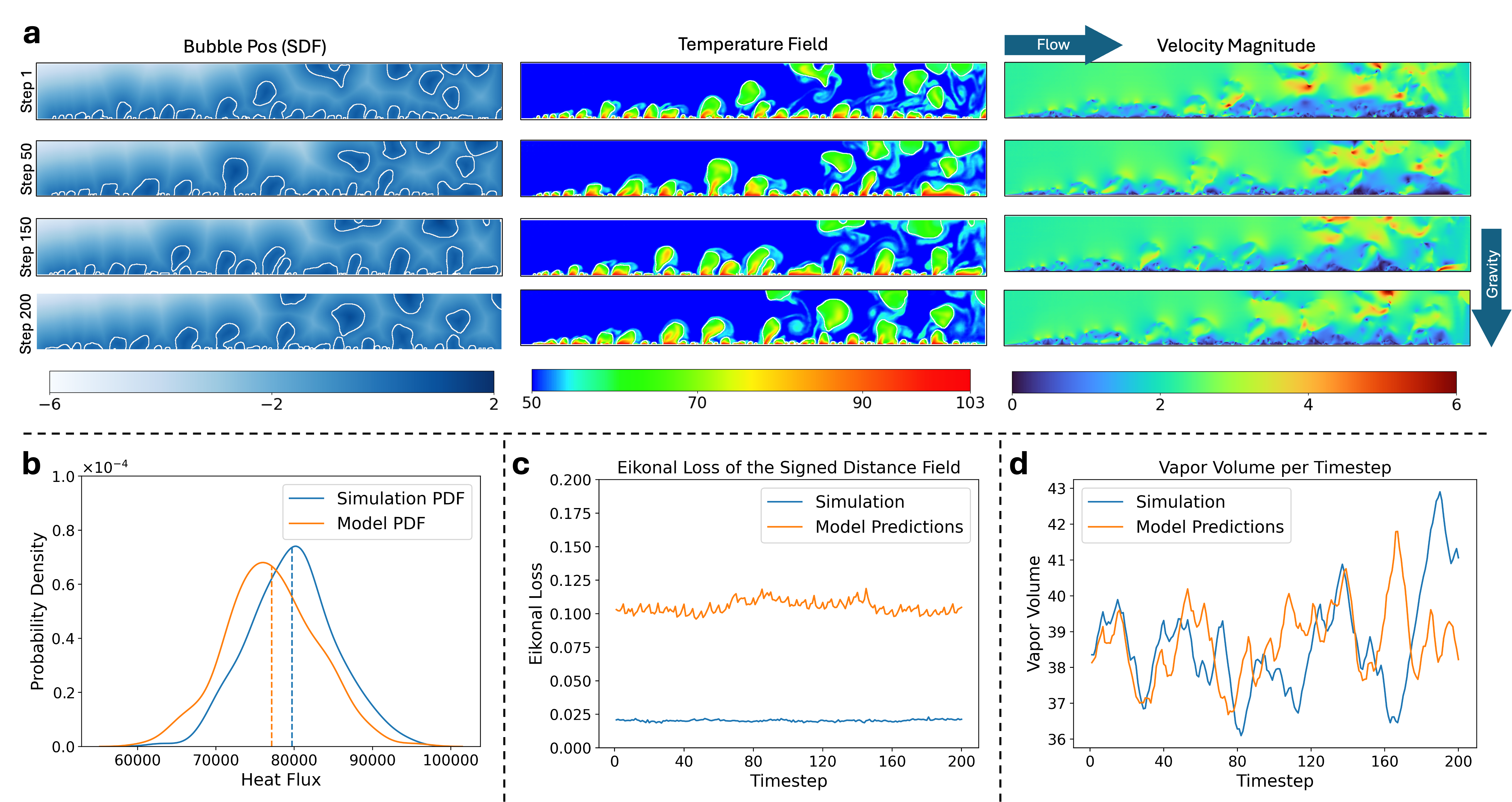}
\caption{\textbf{Flow Boiling Forecasting.} (a) Rollout for a Bubbleformer-S model on an unseen flow boiling trajectory (FC-72, input flow scale = 2.2). (b) Comparison of predicted vs. ground-truth heat flux PDFs. (c) Per-frame Eikonal loss of forecasted signed distance fields  representing bubble positions. (d) Vapor volume in the boiling domain over time for both model and simulation.}
\label{fig:bf-fbvelforecasting}
\end{figure}

We observe that the re-nucleation process learned by Bubbleformer is stochastic: predictions initially align closely with simulations, but diverge gradually over time due to the chaotic nature of the system. 
Despite this divergence, the model's forecasts continue to respect physical laws.
As shown in Figure \ref{fig:bf-fbvelforecasting}, over a 200-step rollout on an unseen flow boiling trajectory, Bubbleformer
predicted fields remain physically well-behaved.
The system level quantities are conserved: the predicted heat flux distribution remain close to that of the simulation, the bubble positions are valid signed distance fields, and mass conservation is closely followed. 
These results show that Bubbleformer does not merely replicate a simulation trajectory, but learns to generate valid realizations of the underlying boiling process.  

Additional discussion of all forecasting models are provided in Appendix \ref{app:forecast}. We also validate our proposed metrics on a deterministic single-bubble simulation study in Appendix \ref{appendix:single-bubble}.
    
\subsection{Prediction of Velocity and Temperature}
\begin{figure}[h]
    \centering
    \includegraphics[width=1.0\linewidth]{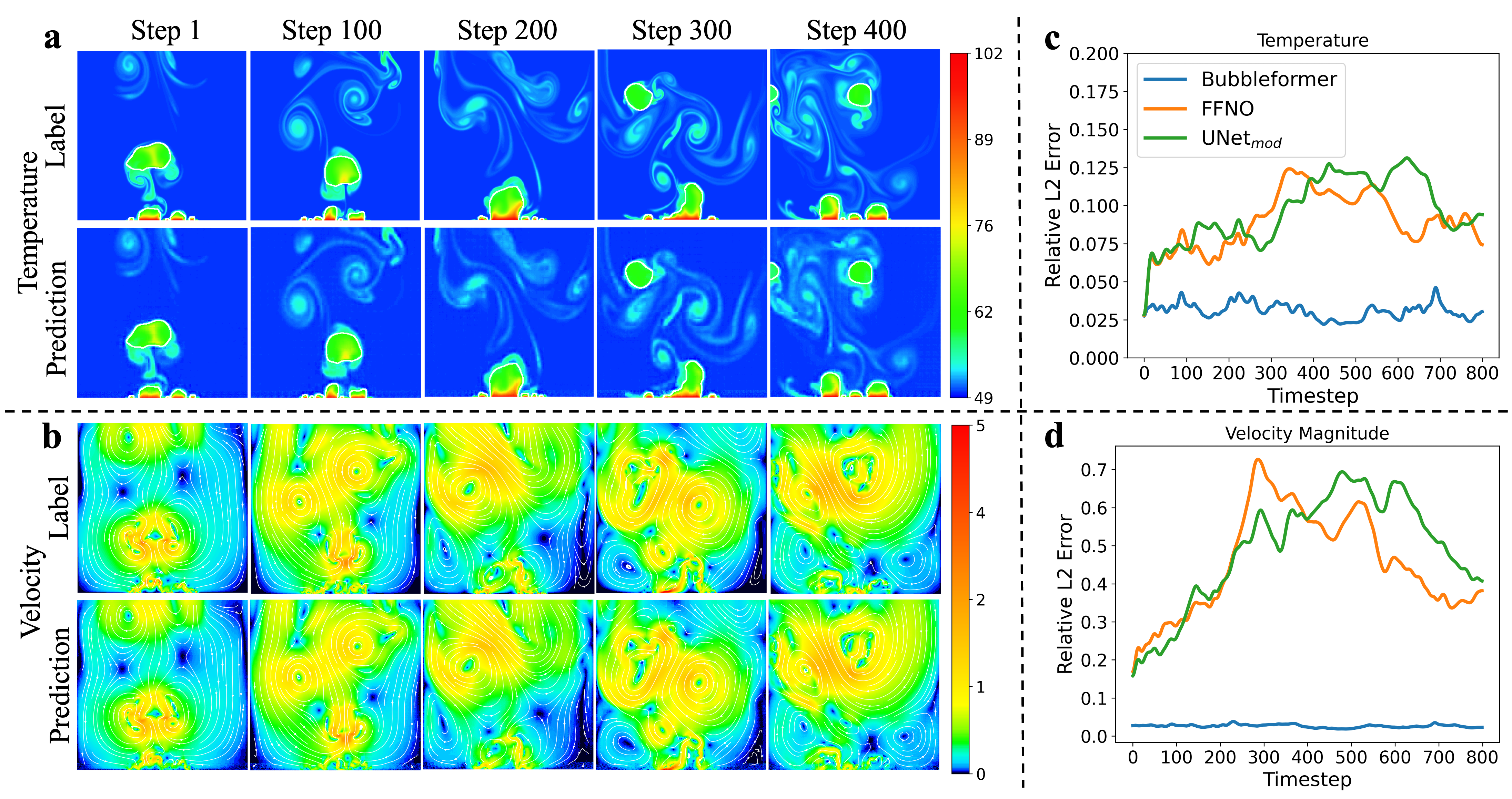}
    \caption{\textbf{Subcooled Pool Boiling Prediction.} 
    (a) and (b) Predicted temperature and velocity fields from the Bubbleformer-S model on an unseen subcooled pool boiling trajectory for FC-72. 
    (c) and (d) Relative L2 error over 800 rollout timesteps for temperature and velocity magnitude (combined x and y components), comparing Bubbleformer-S, FFNO, and UNet\(_{\text{mod}}\). } 
    \label{fig:bf-prediction}
\end{figure}

We evaluate Bubbleformer on the supervised prediction task introduced in BubbleML \cite{hassan2023bubbleml}, modeling temperature and velocity fields in subcooled pool boiling.
As shown in Figure~\ref{fig:bf-prediction}, Bubbleformer outperforms the best-performing models in BubbleML-- UNet$_{\text{mod}}$ and Factorized Fourier Neural Operator (FFNO)--achieving state-of-the-art accuracy across all reported metrics, including relative L2 error. 
To assess long-horizon prediction stability, we extend the autoregressive rollout from 400 to 800 timesteps--doubling the original evaluation setting in BubbleML. 
Bubbleformer maintains accurate predictions over this extended horizon, while baseline models exhibit growing instability.

We attribute Bubbleformer’s performance to its spatiotemporal attention, which enables it to resolve both sharp, non-smooth interfaces characteristic of boiling flows and long-range dependencies. 
While FFNO and UNet$_{\text{mod}}$ suffer  large error spikes during violent bubble detachment events and exhibit a steadily growing error thereafter, Bubbleformer maintains a uniformly low error across the entire 800-step rollout.
Notably, it remains stable even up to 2000 rollout steps, this domenstrates that our spatialtemporal attention architecture is robust to localized topological changes and yields stable long-horizon predictions.
A complete listing of error metrics for each model and dataset pairing can be found in Appendix \ref{app:predict}. 
\section{Conclusions and Limitations} \label{sec:limitations}

We introduce Bubbleformer, a spatiotemporal transformer for forecasting boiling dynamics across fluids, geometries, and regimes.  
Bubbleformer integrates axial attention, frequency-aware scaling, and fluid-conditioned FiLM layers to jointly predict and forecast velocity, temperature, and interface fields.
Our results show that Bubbleformer outperforms prior models in both short-term accuracy and long-horizon stability, while preserving physical consistency as measured by system-level metrics such as heat flux, Eikonal loss, and vapor mass conservation.  
To support generalization, we introduce BubbleML 2.0--a diverse, high-resolution dataset from over 160 simulations.

\textbf{Limitations.} The current Bubbleformer architecture cannot natively operate on AMR grids, necessary for simulating fluids such as water or large real-world domains. 
Interpolation to uniform grids can introduce numerical errors that can make model training unstable. 
Extending the model to directly support AMR inputs--via hierarchical encoding or point-based attention--is an important direction.

Current models are specialized for one type of physics. Combining datasets from different physics (e.g., subcooled and saturated pool boiling) remains challenging. Incorporating patch-level routing via mixture-of-experts models may improve scalability and generalization.

Training on high-resolution datasets requires a large amount of GPU memory, primarily due to the storage of activations during backpropagation. While activation checkpointing can alleviate memory usage, it increases computation time. Neural operators, with their resolution-invariant properties, offer a potential solution by enabling training at lower resolutions. However, their current formulations struggle to capture the sharp discontinuities associated with nucleation events. Advancing neural operators to handle such features is a promising direction for future research in forecasting boiling dynamics.


\begin{ack}
The authors gratefully acknowledge funding support from the Office of Naval Research
(ONR) MURI under Grant No. N000142412575 with Dr. Mark Spector serving as the program officer. We also thank the Research Cyberinfrastructure Center at the University of California, Irvine for the GPU computing resources on the HPC3 cluster.
\end{ack}

\bibliographystyle{abbrv}
\bibliography{main}

\begin{thebibliography}{10}

\bibitem{arnab2021vivit}
A.~Arnab, M.~Dehghani, G.~Heigold, C.~Sun, M.~Lu{\v{c}}i{\'c}, and C.~Schmid.
\newblock Vivit: A video vision transformer.
\newblock In {\em Proceedings of the IEEE/CVF international conference on computer vision}, pages 6836--6846, 2021.

\bibitem{Basu2005}
N.~Basu, G.~R. Warrier, and V.~K. Dhir.
\newblock Wall heat flux partitioning during subcooled flow boiling: Part 1—model development.
\newblock {\em Journal of Heat Transfer}, 127(2):131--140, 03 2005.

\bibitem{gberta_2021_ICML}
G.~Bertasius, H.~Wang, and L.~Torresani.
\newblock Is space-time attention all you need for video understanding?
\newblock In {\em Proceedings of the International Conference on Machine Learning (ICML)}, July 2021.

\bibitem{Bishop:DeepLearning24}
C.~M. Bishop and H.~Bishop.
\newblock {\em Deep Learning: Foundations and Concepts}.
\newblock Springer, 2024.

\bibitem{brandstetter2022message}
J.~Brandstetter, D.~E. Worrall, and M.~Welling.
\newblock Message passing neural {PDE} solvers.
\newblock In {\em International Conference on Learning Representations}, 2022.

\bibitem{chauhan2025neuraloperatorsstrugglelearn}
P.~Chauhan, S.~E. Choutri, M.~Ghattassi, N.~Masmoudi, and S.~E. Jabari.
\newblock Neural operators struggle to learn complex pdes in pedestrian mobility: Hughes model case study, 2025.

\bibitem{chen2023symbolicdiscoveryoptimizationalgorithms}
X.~Chen, C.~Liang, D.~Huang, E.~Real, K.~Wang, Y.~Liu, H.~Pham, X.~Dong, T.~Luong, C.-J. Hsieh, Y.~Lu, and Q.~V. Le.
\newblock Symbolic discovery of optimization algorithms, 2023.

\bibitem{Han1965}
H.~Chi-Yeh and P.~Griffith.
\newblock The mechanism of heat transfer in nucleate pool boiling—part i: Bubble initiaton, growth and departure.
\newblock {\em International Journal of Heat and Mass Transfer}, 8(6):887--904, 1965.

\bibitem{DEVAHDHANUSH2022122603}
V.~Devahdhanush, I.~Mudawar, H.~K. Nahra, R.~Balasubramaniam, M.~M. Hasan, and J.~R. Mackey.
\newblock Experimental heat transfer results and flow visualization of vertical upflow boiling in earth gravity with subcooled inlet conditions – in preparation for experiments onboard the international space station.
\newblock {\em International Journal of Heat and Mass Transfer}, 188:122603, 2022.

\bibitem{DHRUV2024113122}
A.~Dhruv.
\newblock A vortex damping outflow forcing for multiphase flows with sharp interfacial jumps.
\newblock {\em Journal of Computational Physics}, 511:113122, 2024.

\bibitem{DHRUV2019103099}
A.~Dhruv, E.~Balaras, A.~Riaz, and J.~Kim.
\newblock A formulation for high-fidelity simulations of pool boiling in low gravity.
\newblock {\em International Journal of Multiphase Flow}, 120:103099, 2019.

\bibitem{DHRUV2021121826}
A.~Dhruv, E.~Balaras, A.~Riaz, and J.~Kim.
\newblock An investigation of the gravity effects on pool boiling heat transfer via high-fidelity simulations.
\newblock {\em International Journal of Heat and Mass Transfer}, 180:121826, 2021.

\bibitem{Dhruv_Lab-Notebooks_Flash-X-Development_2025_04}
A.~Dhruv and S.~M.~S. Hassan.
\newblock {Github:Lab-Notebooks/Flash-X-Development: 2025.04}.

\bibitem{dosovitskiy2020vit}
A.~Dosovitskiy, L.~Beyer, A.~Kolesnikov, D.~Weissenborn, X.~Zhai, T.~Unterthiner, M.~Dehghani, M.~Minderer, G.~Heigold, S.~Gelly, J.~Uszkoreit, and N.~Houlsby.
\newblock An image is worth 16x16 words: Transformers for image recognition at scale.
\newblock {\em ICLR}, 2021.

\bibitem{DUBEY2022}
A.~Dubey, K.~Weide, J.~O’Neal, A.~Dhruv, S.~Couch, J.~A. Harris, T.~Klosterman, R.~Jain, J.~Rudi, B.~Messer, M.~Pajkos, J.~Carlson, R.~Chu, M.~Wahib, S.~Chawdhary, P.~M. Ricker, D.~Lee, K.~Antypas, K.~M. Riley, C.~Daley, M.~Ganapathy, F.~X. Timmes, D.~M. Townsley, M.~Vanella, J.~Bachan, P.~M. Rich, S.~Kumar, E.~Endeve, W.~R. Hix, A.~Mezzacappa, and T.~Papatheodore.
\newblock Flash-{X}: A multiphysics simulation software instrument.
\newblock {\em SoftwareX}, 19:101168, 2022.

\bibitem{ganesan2018development}
V.~Ganesan.
\newblock Development of a finite volume general two-phase navier-stokes solver for direct numerical simulations on cut-cells with sharp fixed interface.
\newblock Master's thesis, Purdue University, 2018.

\bibitem{ganesan2021review}
V.~Ganesan, R.~Patel, J.~Hartwig, and I.~Mudawar.
\newblock Review of databases and correlations for saturated flow boiling heat transfer coefficient for cryogens in uniformly heated tubes, and development of new consolidated database and universal correlations.
\newblock {\em International Journal of Heat and Mass Transfer}, 179:121656, 2021.

\bibitem{ganesan2021universal}
V.~Ganesan, R.~Patel, J.~Hartwig, and I.~Mudawar.
\newblock Universal critical heat flux (chf) correlations for cryogenic flow boiling in uniformly heated tubes.
\newblock {\em International Journal of Heat and Mass Transfer}, 166:120678, 2021.

\bibitem{ganesan2022universal}
V.~Ganesan, R.~Patel, J.~Hartwig, and I.~Mudawar.
\newblock Universal correlations for post-chf saturated and superheated flow film boiling heat transfer coefficient, minimum heat flux and rewet temperature for cryogenic fluids in uniformly heated tubes.
\newblock {\em International Journal of Heat and Mass Transfer}, 195:123054, 2022.

\bibitem{ganesan2024development}
V.~Ganesan, R.~Patel, J.~Hartwig, and I.~Mudawar.
\newblock Development of two-phase frictional pressure gradient correlation for saturated cryogenic flow boiling in uniformly heated tubes.
\newblock {\em International Journal of Heat and Mass Transfer}, 220:124901, 2024.

\bibitem{Gibou2007}
F.~Gibou, L.~Chen, D.~Nguyen, and S.~Banerjee.
\newblock {A level set based sharp interface method for the multiphase incompressible Navier-Stokes equations with phase change}.
\newblock {\em Journal of Computational Physics}, 222(2):536--555, 2007.

\bibitem{GILMAN201735}
L.~Gilman and E.~Baglietto.
\newblock A self-consistent, physics-based boiling heat transfer modeling framework for use in computational fluid dynamics.
\newblock {\em International Journal of Multiphase Flow}, 95:35--53, 2017.

\bibitem{gupta2022towards}
J.~K. Gupta and J.~Brandstetter.
\newblock Towards multi-spatiotemporal-scale generalized {PDE} modeling.
\newblock {\em arXiv preprint arXiv:2209.15616}, 2022.

\bibitem{hassan2023bubbleml}
S.~M.~S. Hassan, A.~Feeney, A.~Dhruv, J.~Kim, Y.~Suh, J.~Ryu, Y.~Won, and A.~Chandramowlishwaran.
\newblock Bubbleml: A multiphase multiphysics dataset and benchmarks for machine learning.
\newblock In A.~Oh, T.~Naumann, A.~Globerson, K.~Saenko, M.~Hardt, and S.~Levine, editors, {\em Advances in Neural Information Processing Systems}, volume~36, pages 418--449. Curran Associates, Inc., 2023.

\bibitem{hendrycks2023gaussianerrorlinearunits}
D.~Hendrycks and K.~Gimpel.
\newblock Gaussian error linear units (gelus), 2023.

\bibitem{ho2019axialattentionmultidimensionaltransformers}
J.~Ho, N.~Kalchbrenner, D.~Weissenborn, and T.~Salimans.
\newblock Axial attention in multidimensional transformers, 2019.

\bibitem{huang2019ccnet}
Z.~Huang, X.~Wang, L.~Huang, C.~Huang, Y.~Wei, and W.~Liu.
\newblock Ccnet: Criss-cross attention for semantic segmentation.
\newblock In {\em Proceedings of the IEEE/CVF international conference on computer vision}, pages 603--612, 2019.

\bibitem{inanlu2024unveiling}
M.~J. Inanlu, V.~Ganesan, N.~V. Upot, C.~Wang, Z.~Suo, K.~Fazle~Rabbi, P.~Kabirzadeh, A.~Bakhshi, W.~Fu, T.~S. Thukral, et~al.
\newblock Unveiling the fundamentals of flow boiling heat transfer enhancement on structured surfaces.
\newblock {\em Science Advances}, 10(45):eadp8632, 2024.

\bibitem{incropera2007fundamentals}
F.~P. Incropera, D.~P. DeWitt, T.~L. Bergman, and A.~S. Lavine.
\newblock {\em Fundamentals of Heat and Mass Transfer}.
\newblock John Wiley \& Sons, New York, 6th edition, 2007.

\bibitem{Kang2000}
M.~Kang, R.~P. Fedkiw, and X.-D. Liu.
\newblock A boundary condition capturing method for multiphase incompressible flow.
\newblock {\em Journal of Scientific Computing}, 15(3):323--360, Sep 2000.

\bibitem{khodakarami2025mitigatingspectralbiasneural}
S.~Khodakarami, V.~Oommen, A.~Bora, and G.~E. Karniadakis.
\newblock Mitigating spectral bias in neural operators via high-frequency scaling for physical systems, 2025.

\bibitem{kovachki2023neural}
N.~Kovachki, Z.~Li, B.~Liu, K.~Azizzadenesheli, K.~Bhattacharya, A.~Stuart, and A.~Anandkumar.
\newblock Neural operator: Learning maps between function spaces with applications to pdes.
\newblock {\em Journal of Machine Learning Research}, 24(89):1--97, 2023.

\bibitem{kovachki2021neural}
N.~B. Kovachki, Z.~Li, B.~Liu, K.~Azizzadenesheli, K.~Bhattacharya, A.~M. Stuart, and A.~Anandkumar.
\newblock Neural operator: Learning maps between function spaces.
\newblock {\em CoRR}, abs/2108.08481, 2021.

\bibitem{lanthaler2023nonlinear}
S.~Lanthaler, R.~Molinaro, P.~Hadorn, and S.~Mishra.
\newblock Nonlinear reconstruction for operator learning of {PDE}s with discontinuities.
\newblock In {\em The Eleventh International Conference on Learning Representations}, 2023.

\bibitem{lemmon2018nist}
E.~W. Lemmon, I.~H. Bell, M.~Huber, and M.~McLinden.
\newblock Nist standard reference database 23: reference fluid thermodynamic and transport properties-refprop, version 10.0, national institute of standards and technology.
\newblock {\em Standard Reference Data Program, Gaithersburg}, pages 45--46, 2018.

\bibitem{lhoest-etal-2021-datasets}
Q.~Lhoest, A.~Villanova~del Moral, Y.~Jernite, A.~Thakur, P.~von Platen, S.~Patil, J.~Chaumond, M.~Drame, J.~Plu, L.~Tunstall, J.~Davison, M.~{\v{S}}a{\v{s}}ko, G.~Chhablani, B.~Malik, S.~Brandeis, T.~Le~Scao, V.~Sanh, C.~Xu, N.~Patry, A.~McMillan-Major, P.~Schmid, S.~Gugger, C.~Delangue, T.~Matussi{\`e}re, L.~Debut, S.~Bekman, P.~Cistac, T.~Goehringer, V.~Mustar, F.~Lagunas, A.~Rush, and T.~Wolf.
\newblock Datasets: A community library for natural language processing.
\newblock In {\em Proceedings of the 2021 Conference on Empirical Methods in Natural Language Processing: System Demonstrations}, pages 175--184, Online and Punta Cana, Dominican Republic, Nov. 2021. Association for Computational Linguistics.

\bibitem{li2021fourier}
Z.~Li, N.~B. Kovachki, K.~Azizzadenesheli, B.~liu, K.~Bhattacharya, A.~Stuart, and A.~Anandkumar.
\newblock Fourier neural operator for parametric partial differential equations.
\newblock In {\em International Conference on Learning Representations}, 2021.

\bibitem{liang2018pool}
G.~Liang and I.~Mudawar.
\newblock Pool boiling critical heat flux (chf)--part 1: Review of mechanisms, models, and correlations.
\newblock {\em International Journal of Heat and Mass Transfer}, 117:1352--1367, 2018.

\bibitem{lippe2023modeling}
P.~Lippe, B.~S. Veeling, P.~Perdikaris, R.~E. Turner, and J.~Brandstetter.
\newblock Modeling accurate long rollouts with temporal neural {PDE} solvers.
\newblock In {\em ICML Workshop on New Frontiers in Learning, Control, and Dynamical Systems}, 2023.

\bibitem{mccabe2024multiple}
M.~McCabe, B.~R.-S. Blancard, L.~H. Parker, R.~Ohana, M.~Cranmer, A.~Bietti, M.~Eickenberg, S.~Golkar, G.~Krawezik, F.~Lanusse, M.~Pettee, T.~Tesileanu, K.~Cho, and S.~Ho.
\newblock Multiple physics pretraining for spatiotemporal surrogate models.
\newblock In {\em The Thirty-eighth Annual Conference on Neural Information Processing Systems}, 2024.

\bibitem{mccabe2023towards}
M.~McCabe, P.~Harrington, S.~Subramanian, and J.~Brown.
\newblock Towards stability of autoregressive neural operators.
\newblock {\em arXiv preprint arXiv:2306.10619}, 2023.

\bibitem{Mukherjee2007}
A.~Mukherjee and S.~G. Kandlikar.
\newblock {Numerical study of single bubbles with dynamic contact angle during nucleate pool boiling}.
\newblock {\em International Journal of Heat and Mass Transfer}, 50(1-2):127--138, 2007.

\bibitem{perez2017filmvisualreasoninggeneral}
E.~Perez, F.~Strub, H.~de~Vries, V.~Dumoulin, and A.~C. Courville.
\newblock Film: Visual reasoning with a general conditioning layer.
\newblock In {\em AAAI}, 2018.

\bibitem{raffel2023exploringlimitstransferlearning}
C.~Raffel, N.~Shazeer, A.~Roberts, K.~Lee, S.~Narang, M.~Matena, Y.~Zhou, W.~Li, and P.~J. Liu.
\newblock Exploring the limits of transfer learning with a unified text-to-text transformer, 2023.

\bibitem{rassoulinejad2021deep}
S.~M. Rassoulinejad-Mousavi, F.~Al-Hindawi, T.~Soori, A.~Rokoni, H.~Yoon, H.~Hu, T.~Wu, and Y.~Sun.
\newblock Deep learning strategies for critical heat flux detection in pool boiling.
\newblock {\em Applied Thermal Engineering}, 190:116849, 2021.

\bibitem{SATO2017505}
Y.~Sato and B.~Niceno.
\newblock Nucleate pool boiling simulations using the interface tracking method: Boiling regime from discrete bubble to vapor mushroom region.
\newblock {\em International Journal of Heat and Mass Transfer}, 105:505 -- 524, 2017.

\bibitem{SATO2018876}
Y.~Sato and B.~Niceno.
\newblock Pool boiling simulation using an interface tracking method: From nucleate boiling to film boiling regime through critical heat flux.
\newblock {\em International Journal of Heat and Mass Transfer}, 125:876 -- 890, 2018.

\bibitem{sinha2021deep}
K.~N.~R. Sinha, V.~Kumar, N.~Kumar, A.~Thakur, and R.~Raj.
\newblock Deep learning the sound of boiling for advance prediction of boiling crisis.
\newblock {\em Cell Reports Physical Science}, 2(3), 2021.

\bibitem{touvron2022thingsknowvisiontransformers}
H.~Touvron, M.~Cord, A.~El-Nouby, J.~Verbeek, and H.~Jégou.
\newblock Three things everyone should know about vision transformers, 2022.

\bibitem{tran2023factorized}
A.~Tran, A.~Mathews, L.~Xie, and C.~S. Ong.
\newblock Factorized fourier neural operators.
\newblock In {\em The Eleventh International Conference on Learning Representations}, 2023.

\bibitem{wang2020axial}
H.~Wang, Y.~Zhu, B.~Green, H.~Adam, A.~Yuille, and L.-C. Chen.
\newblock Axial-deeplab: Stand-alone axial-attention for panoptic segmentation.
\newblock In {\em European conference on computer vision}, pages 108--126. Springer, 2020.

\bibitem{wang2022antioversmooth}
P.~Wang, W.~Zheng, T.~Chen, and Z.~Wang.
\newblock Anti-oversmoothing in deep vision transformers via the fourier domain analysis: From theory to practice.
\newblock In {\em International Conference on Learning Representations}, 2022.

\bibitem{wilkinson2016fair}
M.~D. Wilkinson, M.~Dumontier, I.~J. Aalbersberg, G.~Appleton, M.~Axton, A.~Baak, N.~Blomberg, J.-W. Boiten, L.~B. da~Silva~Santos, P.~E. Bourne, et~al.
\newblock The fair guiding principles for scientific data management and stewardship.
\newblock {\em Scientific data}, 3(1):1--9, 2016.

\bibitem{williams1989teacherforcing}
R.~J. Williams and D.~Zipser.
\newblock A learning algorithm for continually running fully recurrent neural networks.
\newblock {\em Neural Computation}, 1(2):270--280, 1989.

\bibitem{Xiao2023}
L.~Xiao, Y.~Zhuang, X.~Wu, J.~Yang, Y.~Lu, Y.~Liu, and X.~Han.
\newblock A review of pool-boiling processes based on bubble-dynamics parameters.
\newblock {\em Applied Sciences}, 13(21), 2023.

\bibitem{NEURIPS2023_2d6336c1}
H.~Yang, Y.~Sun, G.~Sundaramoorthi, and A.~Yezzi.
\newblock Steik: Stabilizing the optimization of neural signed distance functions and finer shape representation.
\newblock In A.~Oh, T.~Naumann, A.~Globerson, K.~Saenko, M.~Hardt, and S.~Levine, editors, {\em Advances in Neural Information Processing Systems}, volume~36, pages 13993--14004. Curran Associates, Inc., 2023.

\bibitem{Yazdani2016}
M.~Yazdani, T.~Radcliff, M.~Soteriou, and A.~A. Alahyari.
\newblock {A high-fidelity approach towards simulation of pool boiling}.
\newblock {\em Physics of Fluids}, 28(1):1--30, 2016.

\bibitem{zeng1993unified}
L.~Zeng, J.~Klausner, and R.~Mei.
\newblock A unified model for the prediction of bubble detachment diameters in boiling systems—i. pool boiling.
\newblock {\em International Journal of Heat and Mass Transfer}, 36(9):2261--2270, 1993.

\bibitem{zhang2024mateymultiscaleadaptivefoundation}
P.~Zhang, M.~P. Laiu, M.~Norman, D.~Stefanski, and J.~Gounley.
\newblock Matey: multiscale adaptive foundation models for spatiotemporal physical systems, 2024.

\bibitem{zhao2020prediction}
X.~Zhao, K.~Shirvan, R.~K. Salko, and F.~Guo.
\newblock On the prediction of critical heat flux using a physics-informed machine learning-aided framework.
\newblock {\em Applied Thermal Engineering}, 164:114540, 2020.

\end{thebibliography}

\clearpage

\appendix
\section{Numerical Formulation} \label{app:numerical}

Boiling occurs when a liquid undergoes evaporation on the surface of a solid heater, resulting in formation of gas (vapor) bubbles which induce turbulence and improve heat transfer efficiency. The dynamics of the bubbles is governed by the balance f forces: gravity (buoyancy) pulling the dense liquid downward and pushing vapor upward, surface tension ($\sigma$) trying to minimize the liquid–vapor interface area, and evaporative heat flux. In our numerical model, the liquid-gas interface $\Gamma$ is tracked using a level-set $\phi$, a signed distance function that is positive inside the gas and negative inside the liquid. $\phi = 0$, represents the implicit location of $\Gamma$.

We solve a coupled system of incompressible Navier–Stokes equations and energy equations in each phase. Assuming an incompressible flow, the governing equations can be described for each phase using tensor notations for a 3D domain. 

The equations for \emph{liquid phase} (subscript $L$) are written as:
\begin{subequations} \label{eq:transport-liq}
\begin{equation} \label{eq:momt-liq}
\frac{\partial \vec{\mathbf{u}}}{\partial t} + (\vec{\mathbf{u}}\cdot\nabla)\vec{\mathbf{u}} = -\frac{1}{\rho'_L}\,\nabla P \;+\; \frac{1}{\text{Re}}\,\nabla\cdot\!\Big(\frac{\mu'_L}{\rho'_L}\,\nabla \vec{\mathbf{u}}\Big)\;+\;\frac{\vec{\mathbf{g}}}{\text{Fr}^2}
\end{equation}

\begin{equation} \label{eq:temp-liq}
\frac{\partial T}{\partial t} + \vec{\mathbf{u}}\cdot\nabla T = \frac{1}{\rho'_L\,C'_{pL}\,\text{Pe}}\;\nabla \cdot \Big(k'_L\,\nabla T\Big)\
\end{equation}
\end{subequations}
For the \emph{gas phase} (subscript $G$), the form is analogous:
\begin{subequations} \label{eq:transport-gas}
\begin{equation} \label{eq:momt-gas}
\frac{\partial \vec{\mathbf{u}}}{\partial t} + (\vec{\mathbf{u}}\cdot\nabla)\vec{\mathbf{u}} = -\frac{1}{\rho'_G}\,\nabla P \;+\; \frac{1}{\text{Re}}\,\nabla\cdot\!\Big(\frac{\mu'_G}{\rho'_G}\,\nabla \vec{\mathbf{u}}\Big)\;+\;\frac{\vec{\mathbf{g}}}{\text{Fr}^2}
\end{equation}
\begin{equation} \label{eq:temp-gas}
\frac{\partial T}{\partial t} + \vec{\mathbf{u}}\cdot\nabla T = \frac{1}{\rho'_G\,C'_{pG}\,\text{Pe}}\;\nabla \cdot \Big(k'_G\,\nabla T\Big)\
\end{equation}

\end{subequations}

where $\vec{\mathbf{u}}$ is the velocity field, $P$ is the pressure, and $T$ is the temperature everywhere in the domain. 
The equations are non-dimensionalized with reference quantities from the liquid phase resulting in Reynolds number (${\text{Re}}$) $ = \rho_L u_0 l_0/\mu_L$, Prandtl number (${\text{Pr}}$) $ = \mu_L C_{p_L}/k_L$, Froude number (${\text{Fr}}$) $ = u_0/\sqrt{gl_0}$, and Peclet number (${\text{Pe}}$) $ = \text{Re}\:\text{Pr}$. By construction, all non-dimensional liquid reference properties are normalized to 1 and each gas property is expressed relative to its liquid counterpart:
\begin{equation} \label{eq:ratios}
y_L' =  1,  \quad y_G' =  \frac{y_G}{y_L}
\end{equation}
for any property $y \in {\rho, \mu, C_p, k}$ (density, viscosity, specific heat, and thermal conductivity respectively).
$u_0$ and $l_0$ are reference velocity and length scales, and $g$ is acceleration due to gravity. For boiling problems, the reference length scale is set to the capillary length $l_0 = \sqrt{\sigma/(\rho_L-\rho_G) \: g}$ and the velocity scale is the terminal velocity $u_0=\sqrt{g\:l_0}$. For more details on how these parameters are calculated, we refer the reader to Appendix \ref{app:dataset-details}.
The reference temperature scale is given by $(T-T_{bulk})/\Delta T$, where $\Delta T = T_{wall} - T_{bulk}$ (constant wall temperature) or where $\Delta T = T_{ref} - T_{bulk}$ (constant wall heat flux). 

Within each phase, the flow is incompressible (constant density), so any volume change is solely due to phase change at the interface. 
The continuity equation is written as:
\begin{equation} \label{eq:continuity}
\nabla \cdot {\vec{\mathbf{u}}} = -\,\dot m \; \vec n_\Gamma \cdot \nabla  \frac{1}{\rho'} \:\Biggr\vert_{\:\Gamma}
\end{equation}
where {$\vec n = \nabla \phi/\sqrt{|\nabla \phi|}$} is a unit vector normal to the liquid-gas interface, and $\dot m$ is the evaporative mass flux. Equation \ref{eq:continuity} states that $\nabla \cdot {\vec{\mathbf{u}}} = 0$ everywhere except at the interface, where a source term on the right-hand side accounts for the conversion of mass from liquid to vapor or vice versa.

The interfacial mass flux $\dot m$ is determined by the jump in heat flux at the interface (i.e. how much thermal energy is being used to produce phase change). 
We denote by $\vec n \cdot \nabla T_L$ the difference between heat flux from liquid to gas,  and $\vec n \cdot \nabla T_G$ the heat flux gradient from gas to liquid, normal to the liquid-gas interface $\Gamma$. The difference between these heat fluxes determines the mass transfer from one phase to the other. The value of ${\dot m}$ at the interface is calculated using the energy balance:
\begin{equation} \label{eq:evaporation}
\dot m = \frac{\text{St}}{\text{Pe}} \Bigg( {\vec n_\Gamma \cdot k_L' \: \nabla T_L \:\bigr\vert}_{\:\Gamma} - \vec n_\Gamma \cdot k_G' \: {\nabla T_G  \:\bigr\vert}_{\:\Gamma} \Bigg)
\end{equation}
where ${\text{St}}$ is the Stefan number defined by $\text{St} = C_{p_L} \Delta T/Q_l$, with ${Q_l}$ as latent heat of evaporation.

The level-set function representing the interface is computed using the convection equation:
\begin{equation} \label{eq:levelset-convection}
\frac{\partial \phi}{\partial t} + \vec{\mathbf{u}}_\Gamma \cdot \nabla \phi = 0
\end{equation}

where $\vec{\mathbf{u}}_{\Gamma} = \vec{\mathbf{u}} + (\dot m/\rho')\vec n_\Gamma$ is the \emph{interface velocity}. The convection of level-set is accompanied by a selective reinitialization technique \cite{DHRUV2019103099} to mitigate diffusive errors that are generated from numerical discretization of the convection term. This reinitialization step is essential to preserve accuracy of the interface over long simulations, since the advection equation can cause $\phi$ to lose its signed distance property.

The presence of the interface leads to discontinuities (jumps) in certain field variables. These jump conditions are modeled using a Ghost Fluid Method (GFM) \cite{Gibou2007,DHRUV2024113122}.

\textbf{Velocity Jump.} Reorganizing equation for $\vec{\mathbf{u}}_{\Gamma} = \vec{\mathbf{u}} + (\dot m/\rho') \vec n_{\Gamma}$ results in the following expression for jump in velocity normal to $\Gamma$:

\begin{equation} \label{eq:vel-jump}
[\vec{\mathbf{u}}]_\Gamma \;=\; \vec{\mathbf{u}}_G - \vec{\mathbf{u}}_L \;=\; \dot m \;\vec n_\Gamma \Big(\frac{1}{\rho'_G} - \frac{1}{\rho'_L}\Big)\,
\end{equation}

\textbf{Pressure Jump.} The mass flux $\dot m$, surface tension $\sigma$, and viscous stresses contribute towards a similar jump in pressure, where ${\kappa}$ is the interface curvature. The non-dimensional form for pressure jump using ${\text{We}} = \rho_l u_0^2 l_0/\sigma$ can be written as:
\begin{equation} \label{eq:pres-jump}
[P]_{\Gamma} \;=\; P_G-P_L \;=\; \frac{\kappa}{\text{We}} - \Big( \frac{1}{\rho_G'} -  \frac{1}{\rho_L'} \Big)\, \dot m^2
\end{equation}
The effect of viscous jump is assumed to be negligible in this formulation due to the smeared treatment of viscosity near the interface described in \cite{DHRUV2024113122}. This assumption is consistent with the formulation in \cite{Kang2000}.

\textbf{Temperature Continuity.} Finally, the boundary condition for temperature at the liquid-gas interface is given by:
\begin{equation} \label{eq:temp-jump}
T_{\Gamma}  = T_{sat}
\end{equation}
where $T_{sat}$ corresponds to saturation temperature. 

We consider two types of thermal boundary conditions in our simulations: \emph{constant wall temperature} and \emph{constant wall heat flux}. For a constant wall temperature case, the heater is assigned a non-dimensional temperature using a Dirichlet boundary condition:
\begin{equation}
T'_{wall} = \frac{T_{wall} - T_{bulk}}{\Delta T} = 1
\end{equation}
In contrast, for a constant wall heat flux boundary, the wall temperature is determined using a Neumann boundary condition derived from the specified non-dimensional Nusselt number:
\begin{equation}
\text{Nu}_{wall} = \frac{q \: l_0}{\Delta T \: k_L}
\end{equation}
where $q$ is the imposed wall heat flux. These boundary conditions provide the thermal driving force at the solid surface and govern the rate of phase change at the liquid-gas interface.
\section{Additional BubbleML 2.0 Details} \label{app:dataset-details}

\subsection{Dataset URLs and Links}
\label{appendix:bml-links}

\textbf{Code:} Code for training and evaluation of all the benchmark models are available at our \href{https://github.com/HPCForge/bubbleformer}{Bubbleformer GitHub} repository. The code is released under open MIT license.

\textbf{Dataset:} Our dataset is hosted in Hugging Face at \href{https://huggingface.co/datasets/hpcforge/BubbleML_2}{BubbleML 2.0} under the Creative Commons Attribution 4.0 International License. The dataset can be used with either our published Github code or the Hugging Face datasets\cite{lhoest-etal-2021-datasets} python package.

\textbf{Model Weights:}
Weights for all the benchmark models are available in the \href{https://github.com/HPCForge/bubbleformer/tree/main/model-zoo}{model zoo} directory of the same github repository. All relevant benchmark results can be accessed on the same page.

\textbf{DOI:}
The BubbleML 2.0 dataset has a DOI from Huggingface: \href{https://doi.org/10.57967/hf/5594}{10.57967/hf/5594} and can be cited using the following citation:
\begin{verbatim}
@misc{hpcforge_lab_@_uc_irvine_2025,
	author       = { HPCForge Lab @ UC Irvine and Sheikh Md Shakeel Hassan and
                        Xianwei Zou and Akash Dhruv and Vishwanath Ganesan
                        and Aparna Chandramowlishwaran },
	title        = { BubbleML_2 (Revision 2307458) },
	year         = 2025,
	url          = { https://huggingface.co/datasets/hpcforge/BubbleML_2 },
	doi          = { 10.57967/hf/5594 },
	publisher    = { Hugging Face }
}
\end{verbatim}
\textbf{Documentation and Tutorials:}
We provide detailed documentation (as README files) and Jupyter notebook examples in our \href{https://github.com/HPCForge/bubbleformer/tree/main/examples}{GitHub repository} to load our data, train a model, perform inference using a trained model, and visualize results.

\subsection{Maintenance and Long Term Preservation}
\label{appendix:bml-fair}

The authors are committed to maintaining and preserving this dataset. We have closely worked with Hugging Face to ensure the long term availability of the dataset with the added possibility of extending the dataset without running into storage limits.

\textbf{Findable:} All data is stored in Hugging Face. All present and future data will share a global and persistent DOI \href{https://doi.org/10.57967/hf/5594}{10.57967/hf/5594}.

\textbf{Accessible:} All data and descriptive metadata can be downloaded from Hugging Face using their API or Git.

\textbf{Interoperable:} The data is provided in the form of standard HDF5 files that can be read using many common libraries, such as h5py for Python. Our codebase contains well documented code to load the dataset for various downstream tasks. Relevant metadata is stored as json files with the same name as the simulation. We also provide flexibility to the user to load our data using the Hugging Face datasets\cite{lhoest-etal-2021-datasets} library.

\textbf{Reusable:} BubbleML 2.0 is released under the  Creative Commons Attribution 4.0 International License. Bubbleformer codebase is released under the MIT License.

\subsection{Contact Line Dynamics}

In simulations involving multiple interacting vapor bubbles, realistic modeling of the contact line is essential to accurately capture the bubble dynamics and phase change phenomena. Experimental observations indicate that the contact angle at the bubble base transitions between two limiting values—the receding and advancing contact angles—depending on the direction of contact line motion.

One prevailing interpretation attributes this behavior to inertial effects induced by phase change near the heated surface. During spontaneous evaporation, momentum generated in the direction of vapor departure causes the contact line to recede. When buoyancy forces begin to influence the bubble, the direction of contact line motion reverses. Inertia, now opposing this motion, leads to a gradual increase in the contact angle until the advancing motion dominates and the bubble base begins to shrink. These inertial effects are primarily governed by wall adhesion forces and significantly impact phenomena such as bubble coalescence, where sudden momentum transfers disturb the contact line equilibrium.

To model these effects, Continuum Surface Force (CSF) approaches incorporate the direct contribution of wall adhesion to the liquid-vapor interface force balance \cite{Yazdani2016,SATO2017505,SATO2018876}. In the case of sharp-interface methods such as level set techniques, the contact line is enforced by prescribing a contact angle boundary condition for the level set function. Inertial effects are thus embedded in this boundary condition by dynamically adjusting the local contact angle based on the near-wall flow.

In our implementation, the instantaneous contact angle $\psi$ is defined as a piecewise function of the near-wall radial velocity $u_{base}$ in the plane of the heater surface:

\begin{equation} \label{eq:contact-angle-eq}
\psi = \begin{cases}
\psi_r,& \text{if } u_{base}< 0 \\
\frac{\psi_a-\psi_r}{u_{lim}}u_{base} + \psi_r, & \text{if } 0 \leq u_{base} \leq u_{lim} \\
\psi_a,& \text{if } u_{base}> u_{lim}\\
\end{cases}
\end{equation}

Here, $\psi_r$ and $\psi_a$ represent the receding and advancing contact angles, respectively; $u_{lim}$ is the limiting velocity beyond which the contact angle saturates. This formulation is adapted from the model proposed by Mukherjee et al. \cite{Mukherjee2007}, with the key distinction that our approach attributes inertial effects solely to the advancing motion of the contact line. We find that an optimal value for $u_{lim}$ lies between $20\%$ and $25\%$ of the characteristic velocity, defined as $u_c$. A receding angle of $\psi_r = \pi/4$ is chosen to maintain a balance between surface tension, inertial, and gravitational forces, given that $\sin(\pi/4) = \cos(\pi/4)$. The advancing angle $\psi_a$ is treated as a function of local wall temperature or heat flux.

\subsection{Nucleation Site Distribution and Dynamics}

The distribution and density of nucleation sites on the heater surface are critical parameters for replicating experimental boiling regimes. As wall superheat increases, the number of active nucleation sites typically rises, leading to transitions between boiling regimes on the heat transfer curve \cite{inanlu2024unveiling}.

Several models exist for estimating nucleation site density, ranging from empirical correlations based on observed data \cite{GILMAN201735} to formulations based on surface roughness characteristics \cite{Yazdani2016}. In our framework, we initialize the nucleation site distribution using experimental estimates of bubble density (in bubbles/mm\textsuperscript{2}). These sites are then spatially assigned using a quasi-random low-discrepancy Halton sequence to avoid artificial clustering while maintaining reproducibility.

As wall superheat or heat flux increases during simulation, additional nucleation sites are introduced while preserving the initial configuration. This approach allows us to match experimentally observed bubble site densities without relying heavily on empirical models.

Once the nucleation map is initialized, it acts as a template for bubble generation. At each timestep, the model checks whether the four cells surrounding a nucleation site are filled with liquid. If so, the site is marked for re-nucleation after a specified waiting time $t^*_{wait}$. If vapor reoccupies any of these cells before the wait period elapses, the re-nucleation flag is reset. Newly nucleated bubbles are assigned an initial radius of $0.1\:l_c$.

This dynamic nucleation model allows for physically realistic bubble generation patterns and facilitates the study of transient effects such as intermittent boiling and coalescence-driven heat transfer enhancement.

\subsection{Bubble Wait Time Modeling}

$t^*_{\text{wait}}$ is a critical parameter in the nucleation site model, representing the delay between successive bubble departures at a given site. Its value governs the frequency of nucleation events and thus directly affects heat transfer rates and bubble interactions in the simulation. The modeling of wait time depends on the thermal boundary condition applied to the wall: constant heat flux or constant wall temperature.

\subsubsection{Wait Time for Constant Heat Flux}

For constant heat flux simulations, the bubble frequency is modeled following the heat flux partitioning approach proposed by Basu et al.~\cite{Basu2005}. The frequency $f$ of bubble departure from a nucleation site is calculated using the relationship:
\begin{equation}
f = \frac{q}{N_d E}
\end{equation}
where, $q$ is the applied wall heat flux, $N_d$ is the nucleation site density (bubbles/mm$^2$), $E$ is the energy removed per bubble, given by:
\begin{equation}
E = \frac{4\pi}{3} \rho_v h_{lv} R_d^3
\end{equation}
with $\rho_v$ being the vapor density, $h_{lv}$ the latent heat of vaporization, and $R_d$ the bubble departure radius.

After non-dimensionalization using the capillary length $l_c$ and characteristic velocity $u_c = \sqrt{g_e l_c}$, the inverse bubble frequency becomes:
\begin{equation}
\frac{1}{f} = R_d^3 \cdot \left( \frac{4\pi N_d}{3} \right) \cdot \frac{\rho^* \, \mathrm{Re} \, \mathrm{Pr}}{\mathrm{St} \cdot \mathrm{Nu}_{\text{wall}}}
\end{equation}

Here, $R_d$ is non-dimensional and computed using an empirical relation adapted from Han and Griffith~\cite{Han1965}:
\begin{equation}
R_d = 0.4251 \, \psi \, \sqrt{2\, \mathrm{Bo}}
\end{equation}
where $\psi$ is the static contact angle in radians and $\mathrm{Bo}$ is the Bond number based on $l_c$.

The total bubble life cycle is partitioned into growth and waiting phases, based on empirical assumptions from Xiao et al.~\cite{Xiao2023}:
\begin{equation}
t^*_{\text{growth}} = \frac{1}{4f}, \qquad
t^*_{\text{wait}} = \frac{3}{4f}
\end{equation}

This approach ensures that the nucleation frequency aligns with both energy removal and bubble dynamics under steady heat flux conditions. A Python script to perform this apriori calculation is available in our lab-notebooks \cite{Dhruv_Lab-Notebooks_Flash-X-Development_2025_04}.

\subsubsection{Wait Time for Constant Wall Temperature}

For simulations with a prescribed wall temperature (Dirichlet condition), the bubble dynamics are not directly driven by an imposed heat flux. Instead, we estimate the characteristic time for re-nucleation based on the time a bubble requires to traverse a distance equal to the capillary length $l_c$ at terminal velocity $u_c$. This time scale,
\begin{equation}
t_c = \frac{l_c}{u_c},
\end{equation}
represents the dominant dynamical time for bubble departure under gravity and surface tension balance. Accordingly, the nucleation wait time for constant wall temperature cases is selected to be:
\begin{equation}
t^*_{\text{wait}} \in [0.4,\,0.6, 1.0] \, t_c
\end{equation}
for FC-72, R515B and LN2 respectively. This range accounts for slight variations in bubble terminal dynamics due to interactions or residual vapor near the wall, while maintaining consistency with the overall phase change timescale.

\subsection{Fluids}
We can broadly classify the fluids into cryogens, refrigerants, fluorochemicals, and water based on saturation temperature. Depending on the boiling point, a cryogen is utilized for space applications and cooling superconducting motors, whereas low-temperature refrigerants are used in evaporators and heat exchangers in HVAC\&R industries. Similarly, to maintain the operational temperature of ultra-high heat flux chips at nominal temperature ranges in HPC  datacenters, room temperature fluids such as dielectrics and water are used for direct immersion-based and indirect cold plate-based thermal management systems for electronic cooling. Due to fluid physics unique to these distinct classes of working fluids, these fluids also exhibit distinct thermophysical properties, leading to distinct boiling heat transfer and two-phase flow behavior \cite{ganesan2021review,ganesan2021universal,ganesan2024development}. 

BubbleML 1.0 dataset includes one dielectric fluid. However, to account for distinct fluid physics unique to cryogens, refrigerants, fluorochemicals, and water, in 2.0, we expand the dataset to other working fluids and include at least one from each class of working fluid to cover a wide range of thermodynamic and fluid properties summarized in Table \ref{tab:fluid_properties}.

\begin{table}[htbp]
\centering
\small
\begin{tabular}{l|l|c|c|c|c}
\hline
\textbf{Parameters} & \textbf{Units} & \textbf{Water} & \textbf{FC-72} & \textbf{R515b} & \textbf{LN\textsubscript{2}} \\
\hline
\hline
Saturation Temperature ($T_{sat}$) & °C & 100 & 58 & -19 & -196 \\

Liquid Density ($\rho_l$) & kg·m\textsuperscript{-3} & 958.35 & 1575.6 & 1313.7 & 807 \\

Vapor Density ($\rho_v$) & kg·m\textsuperscript{-3} & 0.5982 & 13.687 & 5.8361 & 4.51 \\

Liquid Viscosity ($\mu_l$) & N·s·m\textsuperscript{-2} & 2.82$\times$10\textsuperscript{-4} & 4.18$\times$10\textsuperscript{-4} & 3.427$\times$10\textsuperscript{-4} & 1.62$\times$10\textsuperscript{-4} \\

Vapor Viscosity ($\mu_v$) & N·s·m\textsuperscript{-2} & 1.232$\times$10\textsuperscript{-5} & 1.177$\times$10\textsuperscript{-5} & 9.626$\times$10\textsuperscript{-6} & 5.428$\times$10\textsuperscript{-6} \\

Liquid Specific Heat Capacity ($C_{p_l}$) & J·kg\textsuperscript{-1}·K\textsuperscript{-1} & 4215.7 & 1099.5 & 1263.6 & 2040.5 \\

Vapor Specific Heat Capacity ($C_{p_v}$) & J·kg\textsuperscript{-1}·K\textsuperscript{-1} & 2080 & 879.30 & 823.26 & 1122.4 \\

Liquid Thermal Conductivity ($k_l$) & W·m\textsuperscript{-1}·K\textsuperscript{-1} & 0.677 & 6.25$\times$10\textsuperscript{-2} & 8.887$\times$10\textsuperscript{-2} & 0.145 \\

Vapor Thermal Conductivity ($k_v$) & W·m\textsuperscript{-1}·K\textsuperscript{-1} & 2.457$\times$10\textsuperscript{-2} & 1.306$\times$10\textsuperscript{-2} & 1.029$\times$10\textsuperscript{-2} & 7.163$\times$10\textsuperscript{-3} \\

Latent Heat of Vaporization ($h_{lv}$) & J·kg\textsuperscript{-1} & 2.256$\times$10\textsuperscript{6} & 8.4227$\times$10\textsuperscript{4} & 1.9056$\times$10\textsuperscript{5} & 1.9944$\times$10\textsuperscript{5} \\

Surface Tension ($\sigma$) & N·m\textsuperscript{-1} & 5.891$\times$10\textsuperscript{-2} & 8.112$\times$10\textsuperscript{-3} & 1.499$\times$10\textsuperscript{-2} & 8.926$\times$10\textsuperscript{-3} \\
\hline
\end{tabular}
\caption{Thermophysical properties of different fluids at saturation under 1 atm. \\Sources: \href{https://www.nist.gov/srd/refprop}{NIST Reference Fluid Thermodynamic and Transport Properties Database \cite{lemmon2018nist}}}
\label{tab:fluid_properties}
\end{table}

The non-dimensional parameters for the simulations of different fluids are in Table \ref{tab:simulation_parameters}. The different thermophysical properties (saturation temperature, liquid/vapor density ratio, surface tension, latent heat of vaporization) impact the physics governing the flows under study in interesting ways. For instance, the low saturation temperature of refrigerants and cryogens affects bubble movement in a way different then that for dielectrics, due to flow of cold liquid towards the heater surface, we can see the bubbles slide on the heater surface befor departure. For instance, owing to low surface tension and latent heat of vaporization, cryogens have been shown to exhibit nucleate boiling dominance at relatively lower wall superheats \cite{ganesan2021review}, leading to a large number of smaller-sized bubbles nucleating as opposed to a relatively small number of larger-sized bubbles nucleating for other room temperature fluids. These unique phenomena cause interesting challenges in the learning of data-driven models.

\begin{table}[htbp]
\centering
\small
\begin{tabular}{Sl|Sc|Sc|Sc|Sc|Sc}
\hline
\textbf{Simulation Parameter} & \textbf{Formula} & \textbf{Water} & \textbf{FC-72} & \textbf{R515b} & \textbf{LN\textsubscript{2}} \\
\hline
\hline
Characteristic Length ($l_c$) (mm) & $\sqrt{\frac{\sigma}{(\rho_l - \rho_v)g}}$ & 2.5 & 0.73 & 1.08 & 1.06 \\

Characteristic Velocity ($u_c$)  (m·s\textsuperscript{-1}) & $\sqrt{gl_c}$ & 0.16 & 0.08 & 0.1 & 0.1 \\

Characteristic Time ($t_c$) (ms) & $\frac{l_c}{u_c}$ & 16 & 8.6 & 10.5 & 10.4 \\

Density ($\rho'$) & $\frac{\rho_v}{\rho_l}$ & 6.242$\times$10\textsuperscript{-4} & 8.687$\times$10\textsuperscript{-3} & 4.442$\times$10\textsuperscript{-3} & 5.589$\times$10\textsuperscript{-3} \\

Viscosity ($\mu'$) & $\frac{\mu_v}{\mu_l}$ & 4.369$\times$10\textsuperscript{-2} & 2.816$\times$10\textsuperscript{-2} & 2.809$\times$10\textsuperscript{-2} & 3.351$\times$10\textsuperscript{-2} \\

Thermal Conductivity ($k'$) & $\frac{k_v}{k_l}$ & 3.629$\times$10\textsuperscript{-2} & 2.09$\times$10\textsuperscript{-1} & 1.158$\times$10\textsuperscript{-1} & 4.94$\times$10\textsuperscript{-2} \\

Specific Heat ($C_p'$) & $\frac{C_{pv}}{C_{pl}}$ & 4.934$\times$10\textsuperscript{-1} & 7.997$\times$10\textsuperscript{-1} & 6.515$\times$10\textsuperscript{-1} & 5.501$\times$10\textsuperscript{-1} \\

Reynolds Number (Re) & $\frac{\rho_l u_c l_c}{\mu_l}$ & 1334 & 231.72 & 426.67 & 542.13 \\

Weber Number (We) & $\frac{\rho_l u_c^2 l_c}{\sigma}$ & 1.0 & 1.0 & 1.0 & 1.0 \\

Prandtl Number (Pr) & $\frac{\mu_l C_{pl}}{k_l}$ & 1.756 & 7.35 & 4.87 & 2.28 \\

Stefan Number (St) & $\frac{C_{pl}}{h_{lv}}\Delta T$ & 0.0019$\Delta T$& 0.013$\Delta T$ & 0.0066$\Delta T$ & 0.0102$\Delta T$ \\
\hline
\end{tabular}
\caption{Non-dimensional simulation parameters for different fluids at saturation under 1 atm. $\Delta T$ is the difference between the maximum and minimum temperatures of the system (heater temperature and bulk liquid temperature)}
\label{tab:simulation_parameters}
\end{table}

For simulation of subcooled and saturated pool boiling of different fluids, we keep ($T_{heater} - T_{bulk}$) consistent across fluids. The bubble growth rate usually scales with Jakob number (in our case St number) that scales with $T_w - T_{sat}$ \cite{zeng1993unified}. Hence, at least for pool boiling, by fixing $T_w - T_{sat}$, we can do a parametric analysis on the bubble growth, merging, and departure dynamics across fluids (effect of thermophysical properties) for the same operating pressure (1 atm) but resulting in different wall heat fluxes (controlled variable in an experiment) due to different HTCs. We choose $T_w$ for the fluids to be greater than $T_{w,ONB}$ (wall temperature required for boiling incipience). Also the corresponding wall heat flux ($q$) should obey $q_{ONB} < q < q_{CHF}$, where the former serves as the lower limit for boiling incipience and the latter determines the upper limit for critical heat flux.


\subsection{Dataset Validation} \label{app:validation}

\textbf{Flow Boiling with Constant Heat Flux.}
\begin{figure}[h]
    \centering
    \includegraphics[width=\linewidth]{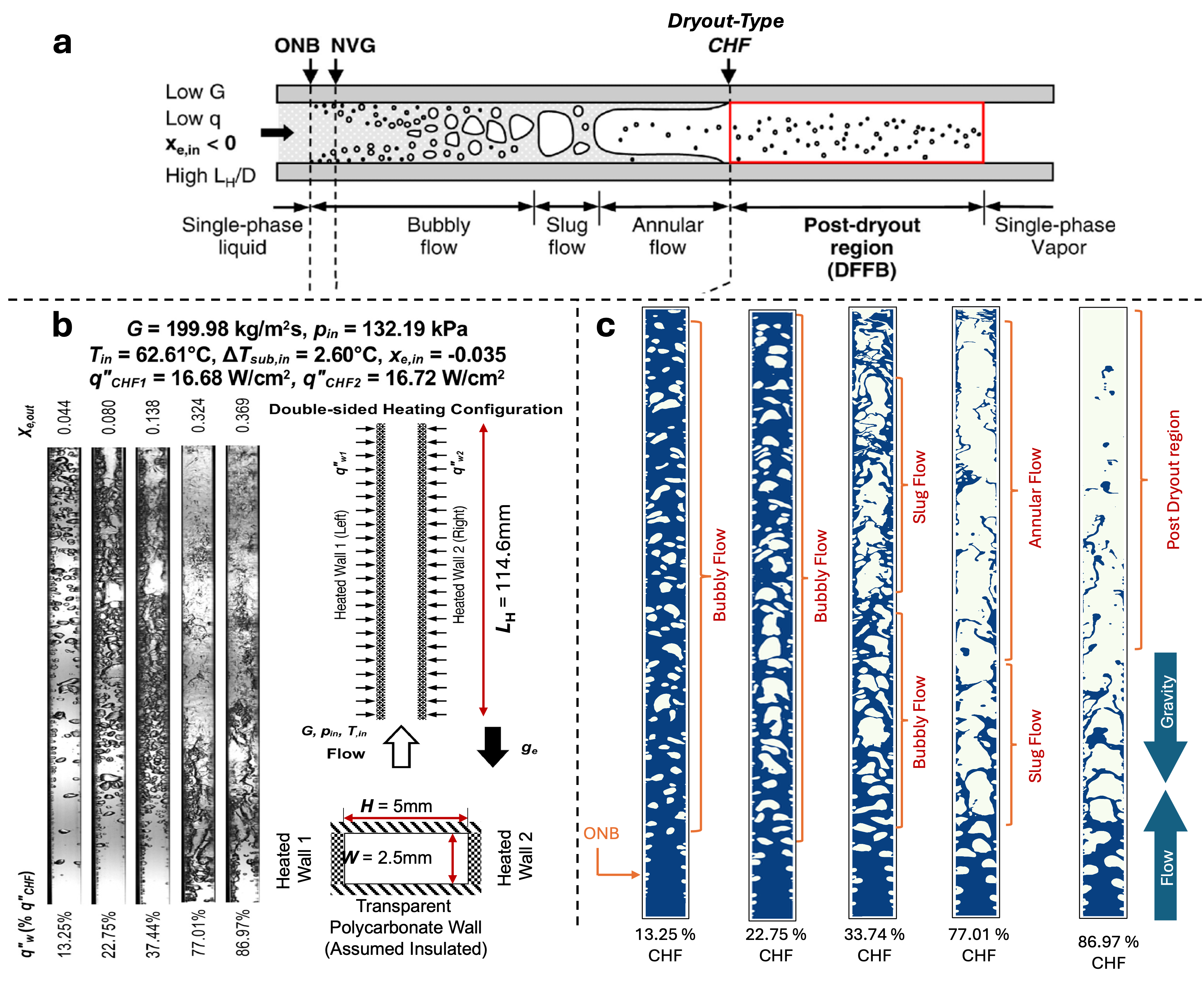}
    \caption{\textbf{Comparison of flow boiling simulations at constant heat flux with experimental observations.} (a) Illustration of a horizontal channel showing different flow boiling regimes~\cite{ganesan2022universal}.
    (b) Experimental visualizations of FC-72 at increasing heat flux values expressed as a fraction of the critical heat flux (CHF)~\cite{DEVAHDHANUSH2022122603}, alongside corresponding 2D simulation domain and boundary conditions. 
    (c) Simulation outputs using Flash-X replicate key features of the flow regimes observed experimentally, including bubbly, slug, annular, and post-dryout flow patterns. 
    Agreement between simulation and experiment validates the fidelity of the numerical framework for modeling flow boiling dynamics.}
    \label{fig:exp-compare}
\end{figure}
To validate flow boiling with a constant heat flux boundary condition, we employed Flash-X to reproduce a parametric study \cite{DEVAHDHANUSH2022122603} of steady-state flow boiling under vertical upflow orientation in a rectangular channel with double-sided heating configuration. The experimental setup used FC-72 as the working fluid, with a saturation temperature of 62°C and a bulk inlet temperature of 60°C, resulting in a moderately subcooled or near-saturated inlet.

Simulations were conducted in a two-dimensional rectangular channel domain measuring 118mm × 5mm, discretized using a block-structured AMR mesh with up to three levels of refinement. This grid configuration ensures adequate resolution of thermal boundary layers and vapor-liquid interfaces while maintaining computational efficiency. Boundary conditions mirrored those in the physical experiment: a velocity-driven inflow on the bottom (x-direction), an outflow on the top, and no-slip walls on all other boundaries. A constant heat flux boundary condition was applied to the walls, set as a percentage of the critical heat flux (CHF).

The fluid properties were defined based on FC-72 specifications. Both liquid and vapor phases were modeled using a multiphase formulation with carefully scaled density and viscosity ratios, as well as other thermophysical parameters, detailed in \cite{Dhruv_Lab-Notebooks_Flash-X-Development_2025_04}.

Gravity was applied in the negative x-direction to reflect the vertical upflow configuration. The simulations used second-order time integration and high-resolution advection via a fifth-order WENO scheme, along with incompressible Navier-Stokes dynamics. A face-centered, divergence-free AMReX interpolator was used to ensure mass conservation across refinement levels. Adaptive mesh refinement (AMR) was triggered using a level-set function to track interface evolution.

Special consideration was given to seeding effects, bubble nucleation timing, and the number of active nucleation sites, which were evaluated a priori and provided as input parameters. While detailed quantitative comparisons of nucleation frequency and bubble dynamics are reserved for future work, the Flash-X simulations show strong qualitative agreement with experimental observations, as illustrated in Figure~\ref{fig:exp-compare}.
\section{Forecasting}
\label{app:forecast}

\subsection{Model Configurations}
We train two Bubbleformer models, details of which are given in Table \ref{tab:model_specs}.
Bubbleformer-S is a smaller variant with reduced embedding and MLP dimensions, while Bubbleformer-L is a larger model with higher capacity. Both models use FiLM conditioning with 9-dimensional thermophysical inputs and axial attention with 12 transformer blocks and 16$\times$16 spatiotemporal patch embeddings.

\begin{table}[!h]
    \centering
    \small
    \begin{tabular}{cccccccccc}
    \toprule
        Model & Embed Dim. & MLP Dim. & FiLM & Heads & Blocks & Patch Size & Params  \\ \midrule
        Bubbleformer-S & 384 & 1536 & 9 & 6 & 12 & 16 & 29.5 M \\
        Bubbleformer-L & 768 & 3072 & 9 & 12 & 12 & 16 & 115.8 M \\
    \bottomrule
    \end{tabular}
    \caption{Architectural specifications of the Bubbleformer models used in our experiments.}
    \label{tab:model_specs}
\end{table}

Each model is trained in a supervised manner using \textbf{teacher forcing} \cite{williams1989teacherforcing} and \textbf{temporal bundling} \cite{brandstetter2022message}. The model inputs are always the simulation ground
truths $[\phi, T, \vec{u}]_{t-k\,:\,t-1}$ and the model learns to predict the next k states, $[\phi, T, \vec{u}]_{t \,:\, t+k-1}$, in a bundled fashion.
We also make the following architectural decisions while training:
\begin{itemize}
    \item \textbf{Patch Embedding and Reconstruction} Hieararchical MLP \cite{touvron2022thingsknowvisiontransformers}.
    \item \textbf{MLP Activation} GeLU \cite{hendrycks2023gaussianerrorlinearunits}.
    \item \textbf{Data Normalization} None. We find that a valid signed distance field is essential to learn renucleation. As such, we do not normalize the data and use a relative L2 loss to perform the learning task.
    \item \textbf{FiLM Layer} 9 fluid parameters are used to condition the model. These parameters are as follows: Reynolds Number(Re), relative specific heat capacity($C_p'$), relative viscosity($\mu'$), relative density($\rho'$), relative thermal conductivity($k'$), Stefan Number(St), Prandtl Number(Pr), nucleation wait time and heater temperature.
    \item \textbf{Attention and Feature Scaling} We use attention scaling in both the spatial and temporal attention blocks, but perform feature scaling only once after each spatio-temporal block. 
\end{itemize}

\textbf{Hardware.} All models were trained using Distributed Data Parallel on 4 NVIDIA A30 GPUs for the Bubbleformer-S models and 2 NVIDIA A100 GPUs for the Bubbleformer-L models. The models were trained for 250 epochs with a single GPU batch size of 4, which took around 48 hours for the S models and 60 hours for the L models.

\subsection{Hyperparameter Settings}

The hyperparameters are tuned for the single bubble validation case and then replicated across all other forecasting scenarios. We found that Lion\cite{chen2023symbolicdiscoveryoptimizationalgorithms} optimizer performs better than Adam and AdamW in our use case. Following the authors recommendations, we set the learning rate to a lower value and weight decay to a higher value than what is generally used for AdamW. We do not perform any data augmentations or transforms during training as we find the existing amount of data to be sufficient for training large transformer models.

\begin{table}[!h]
    \centering
    \caption{\textbf{Training Configuration for Forecasting Experiments.} Summary of hyperparameter settings used during training of Bubbleformer models for boiling trajectory forecasting.}
    \label{tab:hyperparameters-forecast}
    \renewcommand{\arraystretch}{1.2}
    \begin{tabular}{l|c}
        \toprule
        \textbf{Hyperparameter} & \textbf{Value} \\
        \midrule
        Number of Epochs & 250 \\
        Iterations per Epoch & 1000 \\
        Batch Size & 4 \\
        Optimizer & Lion \\
        Weight Decay & 0.1 \\
        Base Learning Rate & $5 \times 10^{-4}$ \\
        Learning Rate Warmup Steps & 1000 \\
        Learning Rate Scheduler & Cosine Annealing \\
        Minimum Learning Rate & $1 \times 10^{-6}$ \\
        History Window Size & 5 \\
        Future Forecast Window & 5 \\
        \bottomrule
    \end{tabular}
\end{table}

\subsection{Single Bubble Validation}
\label{appendix:single-bubble}
Interacting bubble dynamics is a stochastic process. To validate Bubbleformer forecasting, we consider a controlled single bubble originating from a nucleation site. The dynamics include nucleation, growth, departure to the next bubble nucleating at the same location after a wait time. We generate 11 single bubble simulations each for two fluids, FC-72 and R515B corresponding to the same wall superheat ($T_{wall}-T_{liquid}$) values $[29, 32, 33, 34, 36, 37, 38, 40, 41, 42, 45]$. The nucleation wait time is set to different values for the fluids, 0.4 simulation time units for FC-72 and 0.6 simulation time units for R515B. The simulations are run for 400 time units and frames are plotted every 0.2 units to generate 2000 frames per simulation. These simulations are then used to train a Bubbleformer-S forecasting model. The training set consists of 12 simulations corresponding to wall superheat values $[32, 34, 36, 38, 40, 42]$ for both fluids, leaving behind 2 out-of-distribution$[29, 45]$ and 3 in-distribution$[33, 37, 41]$ test trajectories. 

\begin{figure}[h] 
    \centering
    \includegraphics[width=1.0\linewidth]{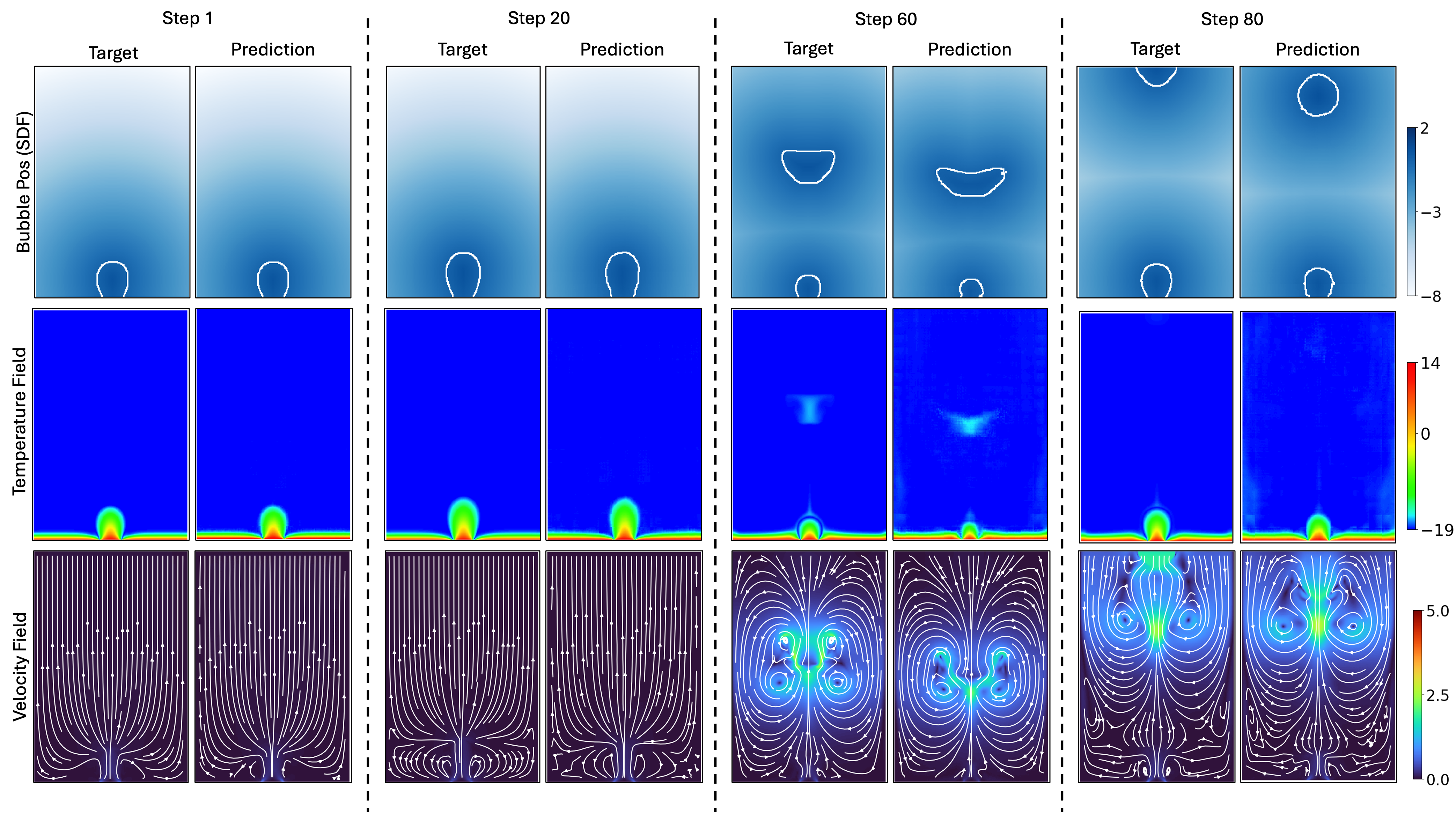} 
    \caption{\textbf{Single-Bubble Forecasting for R515B at 33\,°C Wall Superheat.} Comparison of ground truth and Bubbleformer predicted bubble position (signed distance field), temperature, and velocity fields for an idealized single-bubble scenario. Forecasting was performed using the Bubbleformer-S model under saturated pool boiling conditions.}
    \label{fig:bf-sinbubforecasting_r515b}
\end{figure}

\begin{figure}[h] 
    \centering
    \includegraphics[width=1.0\linewidth]{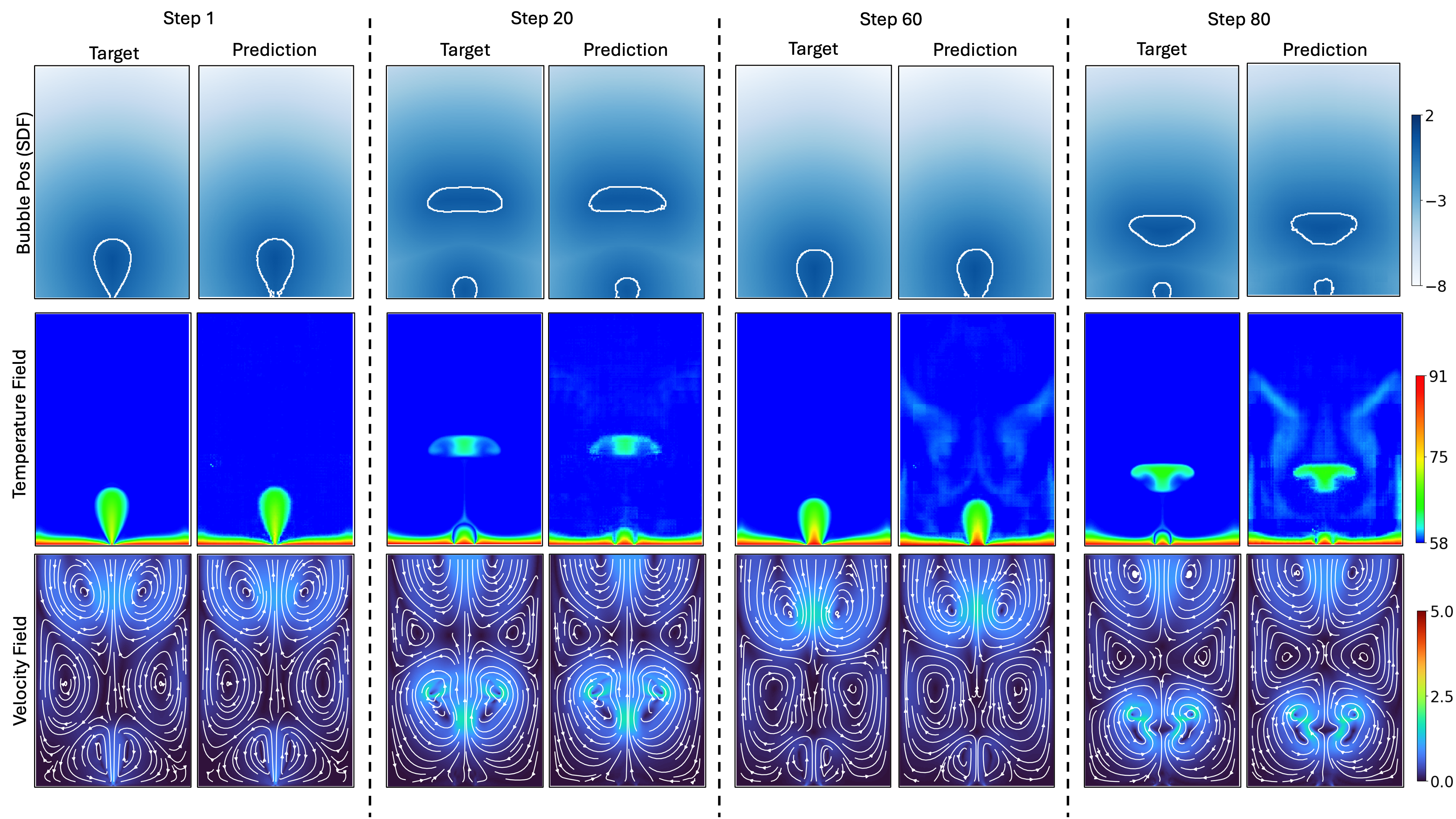} 
    \caption{\textbf{Single-Bubble Forecasting for FC-72 at 33\,°C Wall Superheat.} Comparison of ground truth and Bubbleformer predicted bubble position (signed distance field), temperature, and velocity fields for an idealized single-bubble scenario. Forecasting was performed using the Bubbleformer-S model under saturated pool boiling conditions.}
    \label{fig:bf-sinbubforecasting_fc72}
\end{figure}

\begin{figure}[h] 
    \centering
    \includegraphics[width=1.0\linewidth]{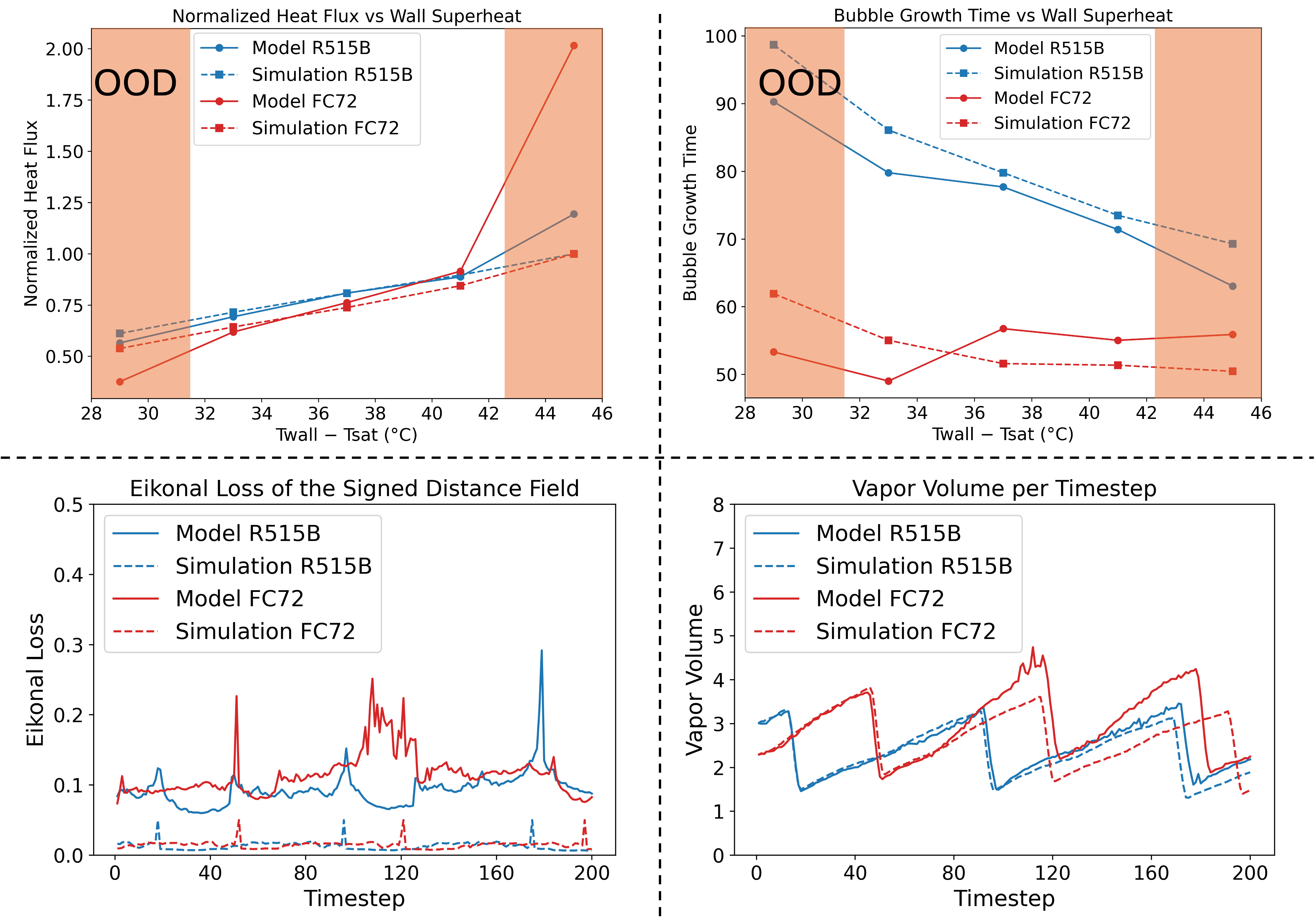} 
    \caption{\textbf{Forecasting Performance: Single Bubble Pool Boiling.} Metrics shown across varying wall superheats and timesteps for R515B and FC-72 under saturated pool boiling conditions. We report bubble growth time(ms), heat flux prediction (system-level quantity), Eikonal loss (interface geometry), and vapor volume (mass conservation) to assess the physical plausibility of forecasts. Results highlight the model’s robustness across fluid types and close alignment with ground truth simulations.}
    \label{fig:bf-sinbubmetrics}
\end{figure}

Upon completion of training, we do an autoregressive rollout for the 10 test trajectories across both fluids for 200 timesteps (around 3 bubble nucleation, growth and departure cycles) and report the physical error metrics for the in-distribution test trajectories in Table \ref{tab:sinbub_error_metrics}. The simulation ground truths and the corresponding model forecasts are compared against each other for one bubble cycle in Figures \ref{fig:bf-sinbubforecasting_r515b} and \ref{fig:bf-sinbubforecasting_fc72}, which shows excellent performance on an in-distribution test trajectory. However, the performance of the model on the out-of-distribution trajectories is significantly worse as seen in Figure \ref{fig:bf-sinbubmetrics} a and b, highlighting a potential failure mode for our Bubbleformer models. We also observe that the results are significantly better for R515B compared to FC-72, especially in the later timesteps of the rollout. While the decreasing trend of bubble growth time with increasing wall superheat is captured correctly for R515B, the model fails to do so for FC-72. Moreover, the vapor volume does not maintain a good correlation with the simulation for the latter half of the rollout in FC-72 which results in incorrect artifacts in the temperature field. This explains the high KL divergence for heat flux seen in FC-72 test trajectories.

\begin{table}[ht]
\centering
\caption{\textbf{Forecasting Accuracy: Single Bubble Pool Boiling.} Physics-based error metrics evaluated at in-distribution wall superheats for the Bubbleformer-S model. Results are reported for two fluids (R515B and FC72) at increasing wall superheat conditions. The metrics quantify accuracy in predicting interface geometry (Eikonal Loss), conserving mass (Relative Vapor Volume [RVV] Error), and capturing heat flux distributions (KL Divergence). Lower values indicate better model performance. }
\label{tab:sinbub_error_metrics}
\renewcommand{\arraystretch}{1.2}
\begin{tabular}{c|ccc|ccc}
\toprule
\textbf{Wall Superheat} & \multicolumn{3}{c|}{\textbf{R515B}} & \multicolumn{3}{c}{\textbf{FC72}} \\
\textbf{(°C)} & \textbf{Eikonal Loss} & \textbf{RVV Err} & \textbf{KL Div} & \textbf{Eikonal Loss} & \textbf{RVV Err} & \textbf{KL Div} \\
\midrule
33 & 0.0755 & 0.1211 & 0.0209 & 0.1468 & 0.2492 & 0.3721 \\
37 & 0.0757 & 0.0534 & 0.0935 & 0.1815 & 0.1765 & 1.062 \\
41 & 0.0769 & 0.0279 & 0.1593 & 0.1541 & 0.1332 & 1.044 \\
\bottomrule
\end{tabular}
\end{table}


\subsection{Pool Boiling Results and Discussion}
The pool boiling dataset consists of 120 simulations spanning across 3 different fluids (FC-72, R515B and LN2) and two different boiling configurations (Saturated and Subcooled Pool Boiling). The simulations span the entire nucleate boiling region of the boiling curve for the fluids, shown in Tables \ref{tab:datasheet-fc72}, \ref{tab:datasheet-r515b}, and \ref{tab:datasheet-ln2}. The learning task is simplified to only two fluids, FC-72 and R515B of a specific boiling configuration. Thus we train 4 forecasting models, a Bubbleformer-S and a Bubbleformer-L each for Saturated and Subcooled pool boiling of the two fluids.

Owing to the poor out-of-distribution performance in the single-bubble study, we leave out 2 in-distribution test trajectories for each fluid to evaluate our trained models. Interestingly, as reported in Table \ref{tab:sat_pb_error_metrics},  we observe that both Bubbleformer-S and Bubbleformer-L perform equivalently well on the saturated pool boiling task. However, in Table \ref{tab:sub_pb_error_metrics}, we observe that for the much harder subcooled pool boiling task, Bubbleformer-L significantly outperforms Bubbleformer-S. We hypothesize that the increased embedding dimension is necessary to represent the richer features such as condensation vortices seen in the subcooled pool boiling study.

\begin{table}[ht]
\centering
\caption{\textbf{Forecasting Accuracy: Saturated Pool Boiling.} Mean error metrics for Bubbleformer-S and Bubbleformer-L models evaluated across two fluids (FC-72 and R515B) and varying wall superheat. The metrics quantify accuracy in predicting interface geometry (Eikonal Loss), conserving mass (Relative Vapor Volume [RVV] Error), and capturing heat flux distributions (KL Divergence). Bolded values indicate the best (lowest) error per fluid-temperature condition.}
\label{tab:sat_pb_error_metrics}
\begin{tabular}{c|c|c|c|c|c}
\toprule
\textbf{Model} & \textbf{Fluid} & \textbf{Heater Temp} & \textbf{Mean Eikonal Loss} & \textbf{Mean RVV Err} & \textbf{KL Div} \\ \midrule
\multirow{4}{*}{Bubbleformer‐S}
  & FC-72  & 91 °C  & 0.132 & 0.073   & \textbf{0.335}         \\ 
  &        & 101 °C & \textbf{0.150} & 0.082   & \textbf{0.277}         \\ 
  & R515B  & 18 °C  & 0.144         & 0.128   & 0.065     \\ 
  &        & 28 °C  & \textbf{0.130}         & 0.094   & 0.145     \\ \midrule
\multirow{4}{*}{Bubbleformer‐L}
  & FC-72  & 91 °C  & \textbf{0.124} & \textbf{0.039}   & 0.360     \\ 
  &        & 101 °C & 0.157          & \textbf{0.042}   & 0.318     \\ 
  & R515B  & 18 °C  & \textbf{0.141}          & \textbf{0.093}   & \textbf{0.023} \\ 
  &        & 28 °C  & 0.137          & \textbf{0.034} & \textbf{0.027}     \\ 
  \bottomrule
\end{tabular}
\end{table}
\begin{table}[ht]
\centering
\caption{\textbf{Forecasting Accuracy: Subcooled Pool Boiling.} Mean error metrics for Bubbleformer-S and Bubbleformer-L models evaluated across two fluids (FC-72 and R515B) and varying wall superheat. The metrics quantify accuracy in predicting interface geometry (Eikonal Loss), conserving mass (Relative Vapor Volume [RVV] Error), and capturing heat flux distributions (KL Divergence). Bolded values indicate the best (lowest) error per fluid-temperature condition.}
\label{tab:sub_pb_error_metrics}
\begin{tabular}{c|c|c|c|c|c}
\toprule
\textbf{Model} & \textbf{Fluid} & \textbf{Heater Temp} & \textbf{Mean Eikonal Loss} & \textbf{Mean RVV Err} & \textbf{KL Div} \\ 
\midrule
\multirow{4}{*}{Bubbleformer‐S} 
  & FC‐72  & 97 °C  & \textbf{0.119} & 2.658 & \textbf{1.060} \\ 
  &        & 107 °C & 0.166 & 2.318 & 0.722 \\ 
  & R515B  & 30 °C  & \textbf{0.136} & 3.656 & 0.154 \\ 
  &        & 40 °C  & \textbf{0.157} & 2.772 & 0.554 \\ 
\midrule
\multirow{4}{*}{Bubbleformer‐L} 
  & FC‐72  & 97 °C  & 0.218 & \textbf{0.376} & 1.407 \\ 
  &        & 107 °C & \textbf{0.163} & \textbf{0.123} & \textbf{0.124} \\ 
  & R515B  & 30 °C  & 0.155 & \textbf{0.078} & \textbf{0.048} \\ 
  &        & 40 °C  & 0.190 & \textbf{0.056} & \textbf{0.049} \\ 
\bottomrule
\end{tabular}
\end{table}

\begin{figure}[h]
    \centering
    \includegraphics[width=1.0\linewidth]{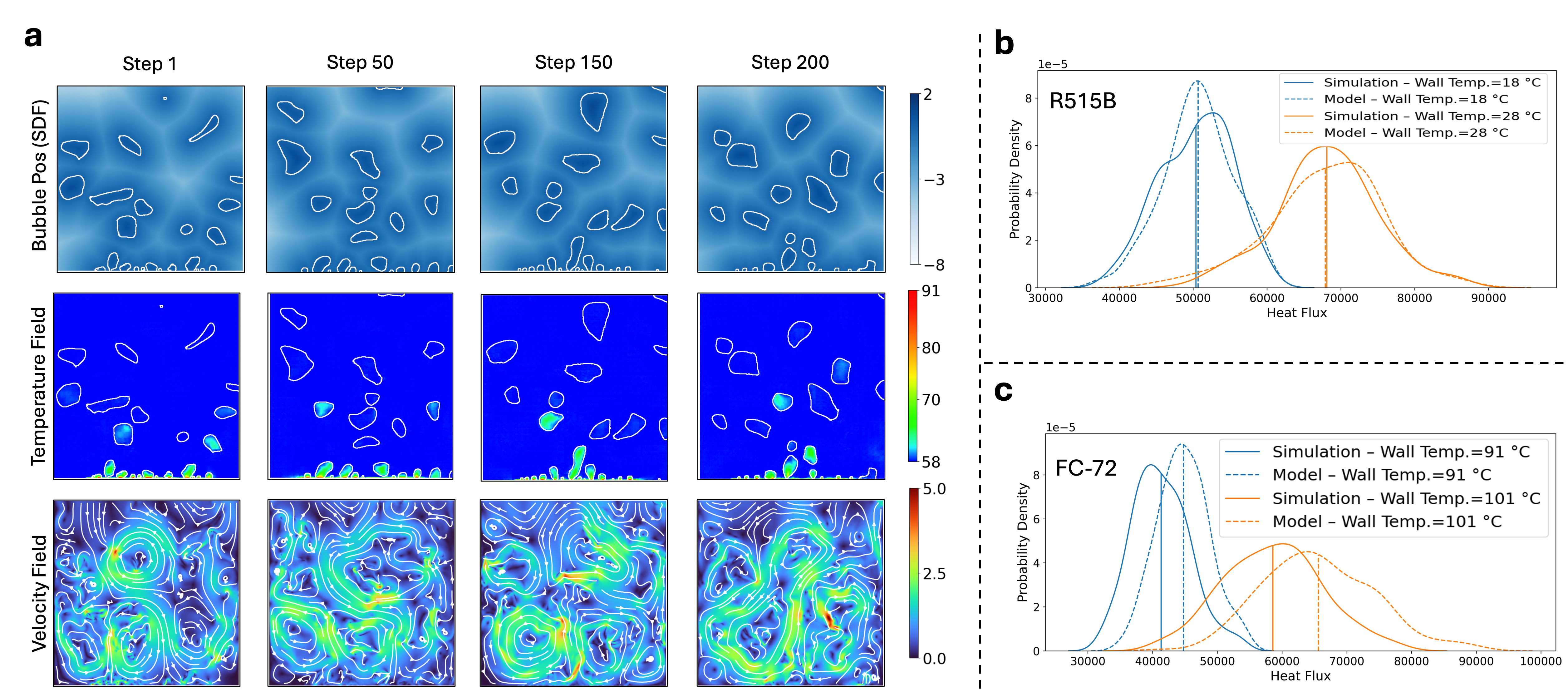}
    \caption{\textbf{Forecasting Results: Saturated Pool Boiling}. (a) Rollout for a Bubbleformer-L model on an unseen pool boiling trajectory, FC-72 at heater temperature= 91 °C. (b) and (c) Comparison of predicted vs. ground-truth heat flux PDFs for R515B and FC-72 respectively at different wall temperatures. }
    \label{fig:bf-pbsatforecasting}
\end{figure}

\begin{figure}[h]
    \centering
    \includegraphics[width=1.0\linewidth]{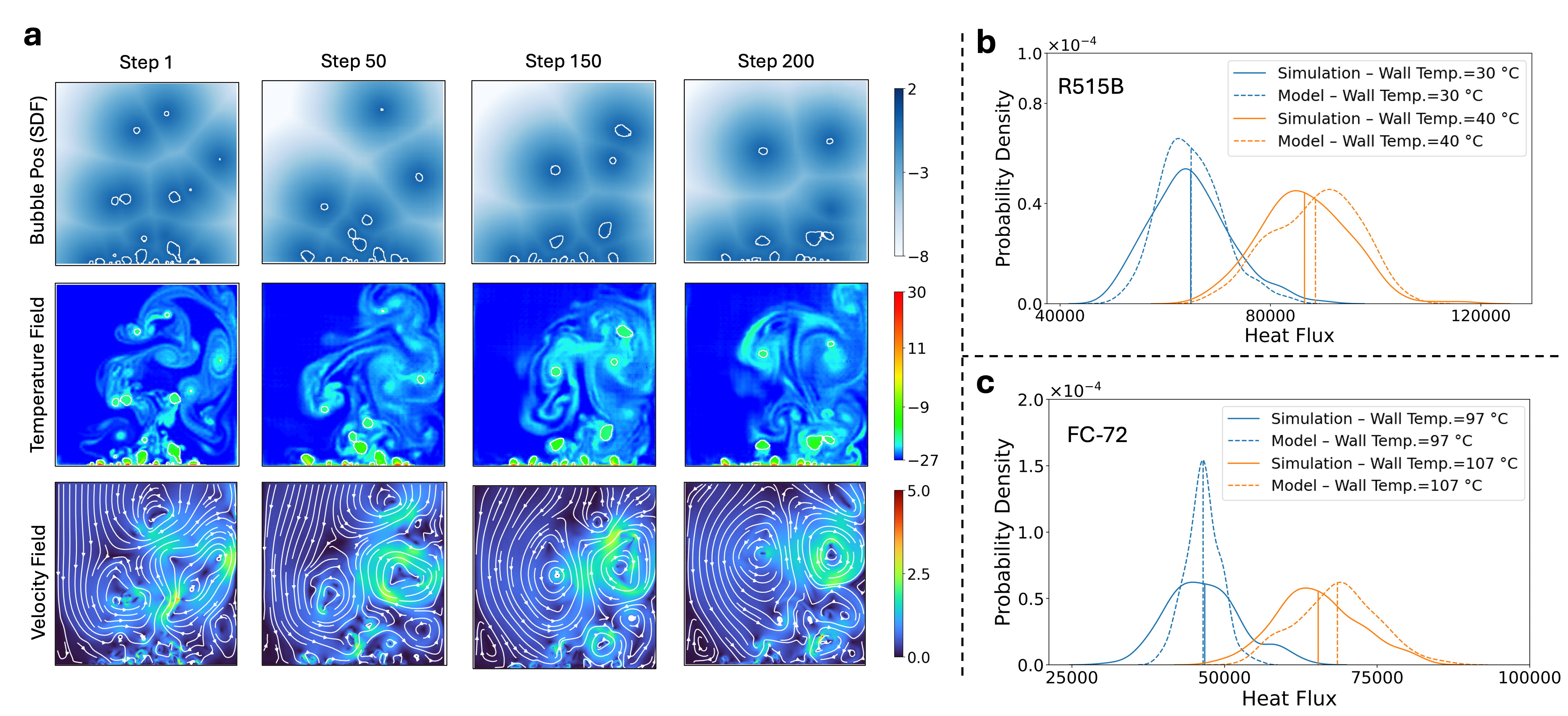}
    \caption{\textbf{Forecasting Results: Subcooled Pool Boiling}. (a) Rollout for a Bubbleformer-L model on an unseen pool boiling trajectory, R515B at heater temperature= 30 °C. (b) and (c) Comparison of predicted vs. ground-truth heat flux PDFs for R515B and FC-72 respectively at different wall temperatures. }
    \label{fig:bf-pbsubforecasting}
\end{figure}

\subsection{Flow Boiling Results and Discussion}

\begin{table}[ht]
\centering
\caption{\textbf{Forecasting Accuracy: Flow Boiling.} Mean error metrics for Bubbleformer-S and Bubbleformer-L on a flow boiling test trajectory with inlet velocity $= 2.2$ m/s using FC-72 as working fluid. Metrics reflect the model’s ability to predict interface geometry (Eikonal Loss), conserve mass (RVV Error), and system-level heat flux distribution (KL Divergence). Lower values indicate better performance.}
\label{tab:error_metrics}
\begin{tabular}{c|c|c|c|c|c}
\toprule
\textbf{Model} & \textbf{Fluid} & \textbf{Inlet Velocity} & \textbf{Mean Eikonal Loss} & \textbf{Mean RVV Err.} & \textbf{KL Div.} \\ 
\midrule
Bubbleformer‐S & FC‐72  & 2.2  & \textbf{0.107} & \textbf{0.105} & \textbf{0.032} \\ 
\midrule
Bubbleformer‐L & FC‐72  & 2.2  & 0.203 & 0.124 & 0.060 \\ 

\bottomrule
\end{tabular}
\end{table}

 Flow boiling forecasting models are trained on the data set with varying inlet velocity scales ranging from 1.5 to 2.9. In this case as well, we leave out trajectory-2.2, an in-distribution test case to evaluate our models. In contrast to UNet and FNO models, the patching mechanism of transformers helps the model learn a global context, which is paramount for learning a good flow boiling forecasting model. However, in this case, we observe that the Bubbleformer-S model performs better than the Bubbleformer-L model across all three metrics. 
\clearpage
\section{Prediction}
\label{app:predict}

\subsection{Benchmark Models} 
Aside from our bubbleformer model, we evaluate two baseline neural PDE solvers used in the BubbleML\cite{hassan2023bubbleml} benchmark: \textbf{UNet\textsubscript{mod}} and \textbf{F-FNO}.

\begin{enumerate}
    \item \textbf{UNet\textsubscript{mod}}: UNet is a commonly used image-to-image architecture in computer vision tasks such as image segmentation. While standard UNet models are not specifically designed for PDE learning—especially when training data come from numerical simulations with varying spatial resolutions—modern adaptations of UNet remain effective in several benchmarks~\cite{gupta2022towards, lippe2023modeling}. UNet\textsubscript{mod} is a variant of UNet that incorporates wide residual connections and group normalization. It is used here as a general-purpose baseline for learning PDE dynamics.

    \item \textbf{F-FNO}: F-FNO (Factorized Fourier Neural Operator)~\cite{tran2023factorized} is a type of neural operator designed to efficiently solve PDEs by learning mappings between function spaces. Neural operators aim to approximate solution operators of PDEs. Given an initial condition \( u_0 \), a neural operator is defined as a mapping \( \mathcal{M} : [0, t_{\text{max}}] \times \mathcal{A} \rightarrow \mathcal{A} \), where \( \mathcal{A} \) is an infinite-dimensional function space, and \( \mathcal{M}(t, u_0) = u_t \)~\cite{li2021fourier, kovachki2023neural}. In practice, training such models requires a large set of initial conditions and their corresponding simulation trajectories, which is computationally expensive for datasets like BubbleML. To mitigate this, we adopt an autoregressive training setup, which has also been used in prior work~\cite{lippe2023modeling, mccabe2023towards, brandstetter2022message}. F-FNO improves the scalability of Fourier Neural Operators by factorizing the Fourier transform across spatial dimensions. This reduces the parameter count per Fourier weight matrix to \( \mathcal{O}(H^2MD) \), enabling the use of more Fourier modes or deeper model architectures. It also introduces a modified residual structure where the residual connection is applied after the nonlinearity.
\end{enumerate}

\subsection{Evaluation Metrics}

To quantitatively assess model performance, we use a set of error metrics computed over spatiotemporal fields. Let $\hat{\mathbf{y}}_t \in \mathbb{R}^{H \times W}$ denote the predicted field at time step $t$, and let $\mathbf{y}_t \in \mathbb{R}^{H \times W}$ be the corresponding ground truth.

\paragraph{Root Mean Square Error (RMSE).}
RMSE measures the average magnitude of the prediction error:
\begin{equation}
    \text{RMSE} = \frac{1}{T} \sum_{t=1}^{T} \sqrt{ \frac{1}{HW} \left\| \hat{\mathbf{y}}_t - \mathbf{y}_t \right\|_2^2 },
\end{equation}
where $T$ is the number of time steps and $H \times W$ denotes the spatial resolution.

\paragraph{Relative L2 Error.}
This metric evaluates the normalized L2 distance between prediction and ground truth:
\begin{equation}
    \text{Relative L2 Error} = \frac{1}{T} \sum_{t=1}^{T} \frac{ \left\| \hat{\mathbf{y}}_t - \mathbf{y}_t \right\|_2 }{ \left\| \mathbf{y}_t \right\|_2 + \varepsilon },
\end{equation}
where $\varepsilon$ is a small constant added for numerical stability.

\paragraph{Max Relative L2 Error.}
We also report the worst-case frame-wise relative error:
\begin{equation}
    \text{Max Relative L2 Error} = \max_{1 \le t \le T} \frac{ \left\| \hat{\mathbf{y}}_t - \mathbf{y}_t \right\|_2 }{ \left\| \mathbf{y}_t \right\|_2 + \varepsilon }.
\end{equation}

\paragraph{Maximum Error.}
This measures the largest pointwise squared error across all spatial and temporal locations:
\begin{equation}
    \text{Max Error} = \max_{t, i, j} \left( \hat{y}_{t, i, j} - y_{t, i, j} \right)^2.
\end{equation}

\paragraph{Interface RMSE.}
We evaluate error specifically on interface regions identified by a signed distance function:
\begin{equation}
    \text{Interface RMSE} = \sqrt{ \frac{1}{|\mathcal{I}|} \sum_{(i,j) \in \mathcal{I}} \left( \hat{y}_{i,j} - y_{i,j} \right)^2 },
\end{equation}
where $\mathcal{I}$ is the set of interface pixel indices.

\paragraph{Boundary RMSE (BRMSE).}
 To assess prediction accuracy near domain edges, we compute RMSE using only boundary values. The boundary is extracted by concatenating the values from all four edges (left, right, top, and bottom) of each frame:
\begin{equation}
    \text{BRMSE} = \sqrt{ \frac{1}{|\mathcal{B}|} \sum_{(i,j) \in \mathcal{B}} \left( \hat{y}_{i,j} - y_{i,j} \right)^2 },
\end{equation}
where $\mathcal{B}$ is the set of all boundary pixels over the entire temporal rollout.

\paragraph{Fourier Spectrum Error.}
To quantify the frequency-dependent discrepancy between predicted and true fields, we compute radial averages of the squared differences in the spatial Fourier domain.

Let $\hat{y}_t, y_t \in \mathbb{R}^{H \times W}$ denote the predicted and true fields at time $t$, and let $\hat{Y}_t = \mathcal{F}[\hat{y}_t]$, $Y_t = \mathcal{F}[y_t]$ be their respective 2D discrete Fourier transforms. Define the Fourier error as:
\begin{equation}
    E_t(i, j) = \left| \hat{Y}_t(i, j) - Y_t(i, j) \right|^2,
\end{equation}
where $(i, j)$ indexes spatial frequencies.

We convert Cartesian frequency coordinates to radial bins via:
\begin{equation}
    k = \left\lfloor \sqrt{i^2 + j^2} \right\rfloor,
\end{equation}
and compute the radially averaged spectrum:
\begin{equation}
    \bar{E}_t(k) = \sum_{\sqrt{i^2 + j^2} \approx k} E_t(i, j).
\end{equation}

We then average across all timesteps:
\begin{equation}
    \bar{E}(k) = \frac{1}{T} \sum_{t=1}^T \bar{E}_t(k),
\end{equation}
normalize by domain size, and report three band-aggregated errors:
\begin{itemize}
    \item \textbf{Low-frequency error:} $\displaystyle \frac{1}{k_\text{low}} \sum_{k=0}^{k_\text{low}-1} \bar{E}(k)$, with $k_\text{low} = 4$
    \item \textbf{Mid-frequency error:} $\displaystyle \frac{1}{k_\text{mid}} \sum_{k=k_\text{low}}^{k_\text{high}-1} \bar{E}(k)$, with $k_\text{mid} = 8$ and $k_\text{high} = 12$
    \item \textbf{High-frequency error:} $\displaystyle \frac{1}{K - k_\text{high}} \sum_{k=k_\text{high}}^{K-1} \bar{E}(k)$, where $K = \min(H/2, W/2)$
\end{itemize}

This metric captures model fidelity across scales: low-$k$ (large structures), mid-$k$ (medium textures), and high-$k$ (fine-grained details). The errors are scaled by the physical domain size $(L_x, L_y)$ to maintain consistency across resolutions.

\subsection{Hyperparameter Settings}

The hyperparameter settings are the same as those used in the forecasting task shown in Table~\ref{tab:hyperparameters-forecast}. The settings for UNet\textsubscript{mod} and F-FNO follow those reported in the BubbleML \cite{hassan2023bubbleml} (Appendix C.3, Table 7).

\subsection{Additional Results and Discussion}

\begin{figure}[h]
    \centering
    \includegraphics[width=1.0\linewidth]{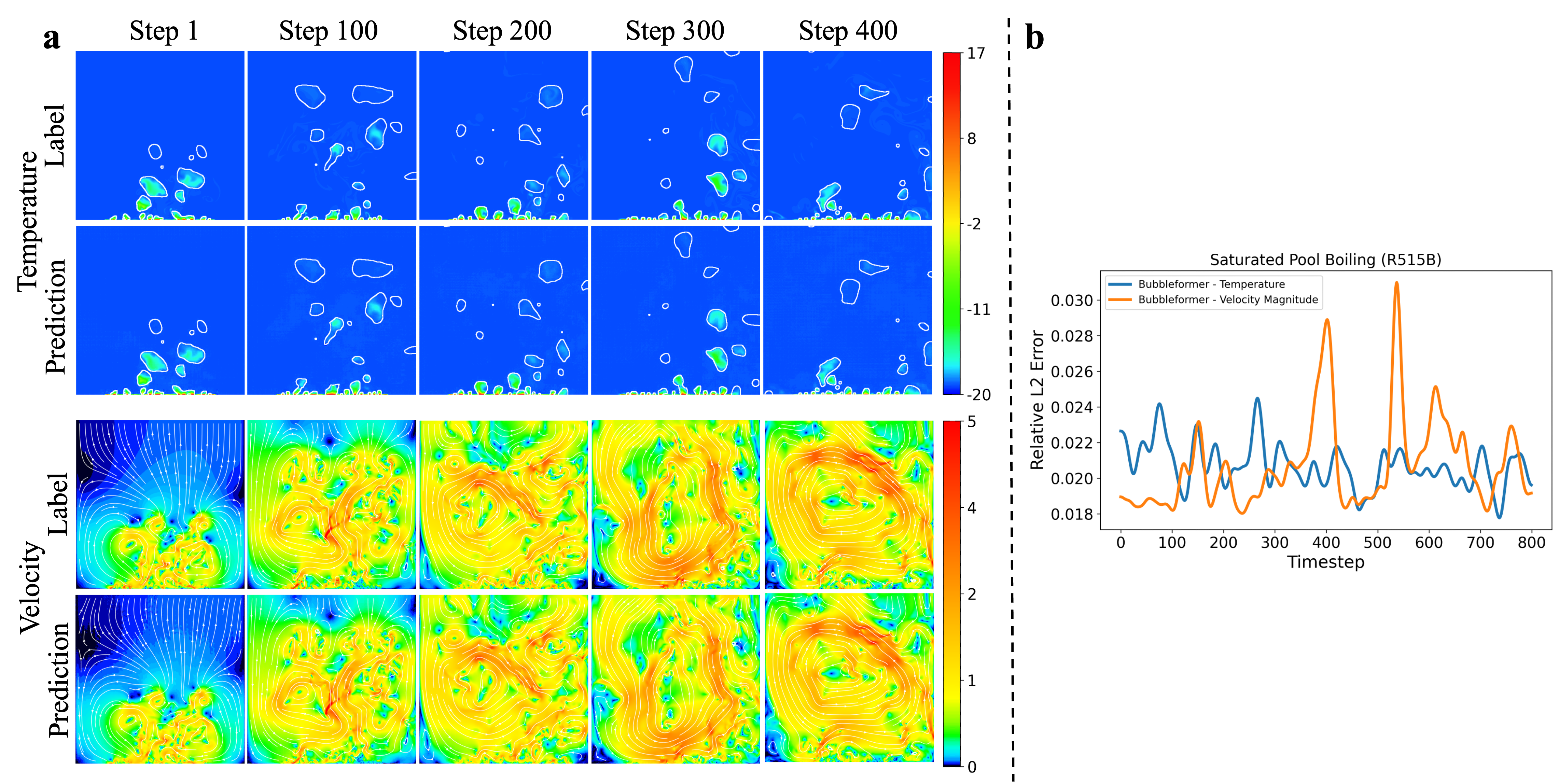}
    \caption{\textbf{Saturated Pool Boiling Prediction.} 
    (a) Predicted temperature and velocity fields from the Bubbleformer-S model on an unseen subcooled pool boiling trajectory for R515B. 
    (b) Relative L2 error over 800 rollout timesteps for temperature and velocity magnitude (combined x and y components) of Bubbleformer-S.} 
    \label{fig:bf-sat-prediction}
\end{figure}

\begin{figure}[h]
    \centering
    \includegraphics[width=1\linewidth]{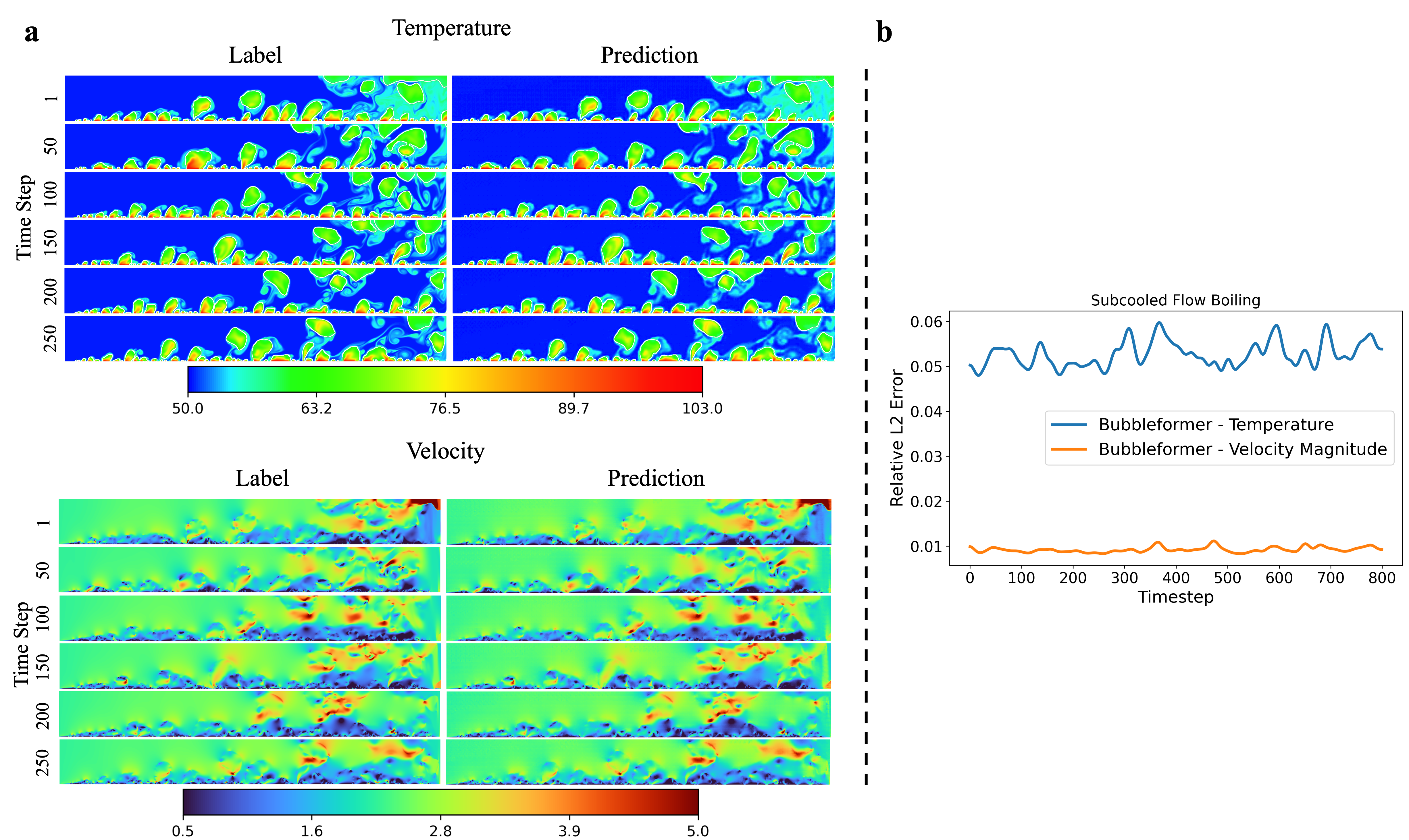}
    \caption{\textbf{Flow Boiling Inlet Velocity Prediction.}  
    (a) Predicted temperature and velocity fields from the Bubbleformer-S model on an unseen flow boiling trajectory for FC-72. 
    (b) Relative L2 error over 800 rollout timesteps for temperature and velocity magnitude (combined x and y components) of Bubbleformer-S.} 
    \label{fig:fb-prediction}
\end{figure}

We present the results for three boiling scenarios in this section: Subcooled Pool Boiling with FC-72 in Table~\ref{tab:bf-metrics-full} and Figure~\ref{fig:bf-prediction}, Saturated Pool Boiling with R-515B in Table~\ref{tab:fb-sat-r515b-metrics} and Figure~\ref{fig:bf-sat-prediction}, and Inlet Velocity Flow Boiling with FC-72 in Table~\ref{tab:fb-metrics} and Figure~\ref{fig:fb-prediction}. For all prediction experiments, we perform rollouts over 800 timesteps.

Additionally, we introduce a max relative L2 error metric, which captures the worst-case mean relative L2 error across the rollout window. 
This metric highlights the model’s robustness under compounding prediction error. 

Bubbleformer outperforms UNet\textsubscript{mod} and F-FNO across most metrics in Table~\ref{tab:bf-metrics-full}, as discussed in Section 6.3. Since UNet\textsubscript{mod} and F-FNO failed to produce valid results for the Flow Boiling cases, we excluded them from Table~\ref{tab:fb-metrics}. Additionally, due to compute and time constraints, we did not evaluate these baselines on the Saturated Pool Boiling scenario in Table~\ref{tab:fb-sat-r515b-metrics}. Nevertheless, we expect Bubbleformer to maintain superior performance, as its inherent ability to resolve both sharp, non-smooth interfaces and long-range dependencies gives it a clear advantage in boiling flow simulations.

\begin{table}[h]
\small
\centering
\begin{tabular}{ll|ccc}
\toprule
\textbf{Category} & \textbf{Metric} & \textbf{Bubbleformer} & \textbf{FFNO} & \textbf{UNet\textsubscript{mod}} \\
\midrule
\multirow{9}{*}{\textbf{Temperature}} & Rel L2 & \textbf{0.031048} & 0.089853 & 0.096906 \\
 & RMSE & \textbf{0.029122} & 0.086781 & 0.093600 \\
 & BRMSE & \textbf{0.109284} & 0.247217 & 0.276180 \\
 & IRMSE & \textbf{0.126722} & 0.210089 & 0.181712 \\
 & MaxErr & 3.675776 & \textbf{2.488442} & 2.790049 \\
 & Max L2 & \textbf{0.075884} & 0.127008 & 0.133258 \\
 & Fourier Low & \textbf{0.180419} & 0.938612 & 0.882603 \\
 & Fourier Mid & \textbf{0.153419} & 0.639546 & 0.758667 \\
 & Fourier High & \textbf{0.051429} & 0.085806 & 0.096905 \\
\midrule
\multirow{9}{*}{\textbf{Velocity X}} & Rel L2 & \textbf{0.173494} & 0.806996 & 0.933271 \\
 & RMSE & \textbf{0.008408} & 0.021742 & 0.024675 \\
 & BRMSE & 0.030632 & 0.023827 & \textbf{0.023260} \\
 & IRMSE & 0.059506 & \textbf{0.050735} & 0.051402 \\
 & MaxErr & 0.785328 & \textbf{0.347980} & 0.369229 \\
 & Max L2 & \textbf{0.405068} & 1.326638 & 1.506975 \\
 & Fourier Low & \textbf{0.049933} & 0.340134 & 0.417801 \\
 & Fourier Mid & \textbf{0.042705} & 0.124967 & 0.125568 \\
 & Fourier High & 0.015219 & 0.011966 & \textbf{0.011803} \\
\midrule
\multirow{9}{*}{\textbf{Velocity Y}} & Rel L2 & \textbf{0.026967} & 0.681400 & 0.765296 \\
 & RMSE & \textbf{0.008006} & 0.024380 & 0.026884 \\
 & BRMSE & \textbf{0.012534} & 0.016269 & 0.018310 \\
 & IRMSE & 0.051713 & 0.048940 & \textbf{0.047879} \\
 & MaxErr & 1.699735 & \textbf{0.791652} & 0.843999 \\
 & Max L2 & \textbf{0.062572} & 1.247504 & 1.315714 \\
 & Fourier Low & \textbf{0.059147} & 0.346206 & 0.373078 \\
 & Fourier Mid & \textbf{0.043992} & 0.134964 & 0.142891 \\
 & Fourier High & 0.013107 & \textbf{0.010859} & 0.011118 \\
\bottomrule
\end{tabular}
\caption{\textbf{Subcooled Pool Boiling (FC-72) Velocity and Temperature Prediction Metrics.} Metrics include Relative L2 error, RMSE, BRMSE, Max Error, Maximum L2 error, and Fourier Errors for each physical field.}
\label{tab:bf-metrics-full}
\end{table}

\begin{table}[h]
\small
\centering
\begin{tabular}{ll|c}
\toprule
\textbf{Category} & \textbf{Metric} & \textbf{Bubbleformer} \\
\midrule
\multirow{9}{*}{\textbf{Temperature}} & Rel L2 & 0.020703 \\
 & RMSE & 0.018814 \\
 & BRMSE & 0.121156 \\
 & MaxErr & 3.107522 \\
 & Max L2 & 0.040473 \\
 & Fourier Low & 0.023166 \\
 & Fourier Mid & 0.041523 \\
 & Fourier High & 0.033939 \\
\midrule
\multirow{9}{*}{\textbf{Velocity X}} & Rel L2 & 0.099593 \\
 & RMSE & 0.004759 \\
 & BRMSE & 0.009329 \\
 & IRMSE & 0.016655 \\
 & MaxErr & 1.969497 \\
 & Max L2 & 0.310537 \\
 & Fourier Low & 0.018420 \\
 & Fourier Mid & 0.016868 \\
 & Fourier High & 0.007180 \\
\midrule
\multirow{9}{*}{\textbf{Velocity Y}} & Rel L2 & 0.021445 \\
 & RMSE & 0.004838 \\
 & BRMSE & 0.004315 \\
 & IRMSE & 0.016664 \\
 & MaxErr & 0.674551 \\
 & Max L2 & 0.057292 \\
 & Fourier Low & 0.019422 \\
 & Fourier Mid & 0.017947 \\
 & Fourier High & 0.007236 \\
\bottomrule
\end{tabular}
\caption{\textbf{Saturated Pool Boiling (R515B) Prediction Metrics}}
\label{tab:fb-sat-r515b-metrics}
\end{table}

\begin{table}[h]
\small
\centering
\begin{tabular}{ll|c}
\toprule
\textbf{Category} & \textbf{Metric} & \textbf{Bubbleformer} \\
\midrule
\multirow{9}{*}{\textbf{Temperature}} & Rel L2 & 0.052712 \\
 & RMSE & 0.046211 \\
 & BRMSE & 0.178124 \\
 & IRMSE & 0.121292 \\
 & MaxErr & 3.139554 \\
 & Max L2 & 0.110760 \\
 & Fourier Low & 0.070464 \\
 & Fourier Mid & 0.120353 \\
 & Fourier High & 0.146381 \\
\midrule
\multirow{9}{*}{\textbf{Velocity X}} & Rel L2 & 0.009240 \\
 & RMSE & 0.004052 \\
 & BRMSE & 0.013071 \\
 & IRMSE & 0.012441 \\
 & MaxErr & 0.432624 \\
 & Max L2 & 0.029023 \\
 & Fourier Low & 0.026810 \\
 & Fourier Mid & 0.018300 \\
 & Fourier High & 0.011747 \\
\midrule
\multirow{9}{*}{\textbf{Velocity Y}} & Rel L2 & 0.306108 \\
 & RMSE & 0.005391 \\
 & BRMSE & 0.008027 \\
 & IRMSE & 0.017001 \\
 & MaxErr & 0.658709 \\
 & Max L2 & 0.598994 \\
 & Fourier Low & 0.011650 \\
 & Fourier Mid & 0.014031 \\
 & Fourier High & 0.016585 \\

\bottomrule
\end{tabular}
\caption{\textbf{Flow Boiling Inlet Velocity (FC-72) Prediction Metrics.}}
\label{tab:fb-metrics}
\end{table}

\clearpage

\section{Datasheet}
\begin{table}[htbp]
\small
\begin{tabular}{l|l|c|c|c|c|c}
\hline
\textbf{Study} & \textbf{Fluid (T\_bulk)} & \textbf{Wall Temp.} & \textbf{Nuc. Sites} & \textbf{T\_wall - T\_bulk} & \textbf{Sat. Temp.} & \textbf{Stefan Num.} \\
\hline
\hline
\multirow{20}{*}{Saturated} & \multirow{20}{*}{FC-72 (58\textdegree C)} & 75 & 5 & 17 & 0 & 0.2219 \\
 & & 76 & 6 & 18 & 0 & 0.2349 \\
 & & 78 & 8 & 20 & 0 & 0.2610 \\
 & & 80 & 10 & 22 & 0 & 0.2871 \\
 & & 82 & 12 & 24 & 0 & 0.3132 \\
 & & 84 & 14 & 26 & 0 & 0.3393 \\
 & & 86 & 16 & 28 & 0 & 0.3654 \\
 & & 88 & 18 & 30 & 0 & 0.3915 \\
 & & 90 & 20 & 32 & 0 & 0.4176 \\
 & & 91 & 21 & 33 & 0 & 0.4307 \\
 & & 92 & 22 & 34 & 0 & 0.4437 \\
 & & 94 & 24 & 36 & 0 & 0.4698 \\
 & & 96 & 26 & 38 & 0 & 0.4959 \\
 & & 98 & 28 & 40 & 0 & 0.5220 \\
 & & 100 & 30 & 42 & 0 & 0.5481 \\
 & & 101 & 31 & 43 & 0 & 0.5612 \\
 & & 102 & 32 & 44 & 0 & 0.5742 \\
 & & 104 & 34 & 46 & 0 & 0.6003 \\
 & & 106 & 36 & 48 & 0 & 0.6264 \\
 & & 107 & 37 & 49 & 0 & 0.6395 \\
\hline
\multirow{20}{*}{Subcooled} & \multirow{20}{*}{FC-72 (50\textdegree C)} & 85 & 3 & 35 & 0.2286 & 0.4568 \\
 & & 86 & 4 & 36 & 0.2222 & 0.4698 \\
 & & 88 & 6 & 38 & 0.2105 & 0.4959 \\
 & & 90 & 8 & 40 & 0.2000 & 0.5220 \\
 & & 92 & 10 & 42 & 0.1905 & 0.5481 \\
 & & 94 & 12 & 44 & 0.1818 & 0.5742 \\
 & & 96 & 14 & 46 & 0.1739 & 0.6003 \\
 & & 97 & 15 & 47 & 0.1702 & 0.6134 \\
 & & 98 & 16 & 48 & 0.1667 & 0.6264 \\
 & & 100 & 18 & 50 & 0.1600 & 0.6525 \\
 & & 102 & 20 & 52 & 0.1538 & 0.6786 \\
 & & 104 & 22 & 54 & 0.1481 & 0.7047 \\
 & & 106 & 24 & 56 & 0.1429 & 0.7308 \\
 & & 107 & 25 & 57 & 0.1404 & 0.7439 \\
 & & 108 & 26 & 58 & 0.1379 & 0.7569 \\
 & & 110 & 28 & 60 & 0.1333 & 0.7830 \\
 & & 112 & 30 & 62 & 0.1290 & 0.8091 \\
 & & 114 & 32 & 64 & 0.1250 & 0.8352 \\
 & & 116 & 34 & 66 & 0.1212 & 0.8613 \\
 & & 117 & 35 & 67 & 0.1194 & 0.8744 \\
\hline
\end{tabular}
\caption{Boiling Curve Data for FC-72}
\label{tab:datasheet-fc72}
\end{table}

\begin{table}[htbp]
\small
\begin{tabular}{l|l|c|c|c|c|c}
\hline
\textbf{Study} & \textbf{Fluid (T\_bulk)} & \textbf{Wall Temp.} & \textbf{Nuc. Sites} & \textbf{T\_wall - T\_bulk} & \textbf{Sat. Temp.} & \textbf{Stefan Num.} \\
\hline
\hline
\multirow{20}{*}{Saturated} & \multirow{20}{*}{R515B (-19\textdegree C)} & 4 & 5 & 23 & 0 & 0.1525 \\
 & & 5 & 6 & 24 & 0 & 0.1591 \\
 & & 7 & 8 & 26 & 0 & 0.1724 \\
 & & 9 & 10 & 28 & 0 & 0.1857 \\
 & & 11 & 12 & 30 & 0 & 0.1989 \\
 & & 13 & 14 & 32 & 0 & 0.2122 \\
 & & 15 & 16 & 34 & 0 & 0.2255 \\
 & & 17 & 18 & 36 & 0 & 0.2387 \\
 & & 18 & 19 & 37 & 0 & 0.2453 \\
 & & 19 & 20 & 38 & 0 & 0.2520 \\
 & & 21 & 22 & 40 & 0 & 0.2652 \\
 & & 23 & 24 & 42 & 0 & 0.2785 \\
 & & 25 & 26 & 44 & 0 & 0.2918 \\
 & & 27 & 28 & 46 & 0 & 0.3050 \\
 & & 28 & 29 & 47 & 0 & 0.3117 \\
 & & 29 & 30 & 48 & 0 & 0.3183 \\
 & & 31 & 32 & 50 & 0 & 0.3316 \\
 & & 33 & 34 & 52 & 0 & 0.3448 \\
 & & 35 & 36 & 54 & 0 & 0.3581 \\
 & & 36 & 37 & 55 & 0 & 0.3647 \\
\hline
\multirow{20}{*}{Subcooled} & \multirow{20}{*}{R515B (-27\textdegree C)} & 14 & 3 & 41 & 0.1951 & 0.2719 \\
 & & 15 & 4 & 42 & 0.1905 & 0.2785 \\
 & & 17 & 6 & 44 & 0.1818 & 0.2918 \\
 & & 19 & 8 & 46 & 0.1739 & 0.3050 \\
 & & 21 & 10 & 48 & 0.1667 & 0.3183 \\
 & & 23 & 12 & 50 & 0.1600 & 0.3316 \\
 & & 25 & 14 & 52 & 0.1538 & 0.3448 \\
 & & 27 & 16 & 54 & 0.1481 & 0.3581 \\
 & & 29 & 18 & 56 & 0.1429 & 0.3713 \\
 & & 30 & 19 & 57 & 0.1404 & 0.3780 \\
 & & 31 & 20 & 58 & 0.1379 & 0.3846 \\
 & & 33 & 22 & 60 & 0.1333 & 0.3979 \\
 & & 35 & 24 & 62 & 0.1290 & 0.4111 \\
 & & 37 & 26 & 64 & 0.1250 & 0.4244 \\
 & & 39 & 28 & 66 & 0.1212 & 0.4376 \\
 & & 40 & 29 & 67 & 0.1194 & 0.4443 \\
 & & 41 & 30 & 68 & 0.1176 & 0.4509 \\
 & & 43 & 32 & 70 & 0.1143 & 0.4642 \\
 & & 45 & 34 & 72 & 0.1111 & 0.4774 \\
 & & 46 & 35 & 73 & 0.1096 & 0.4841 \\
\hline
\end{tabular}
\caption{Boiling Curve Data for R515B}
\label{tab:datasheet-r515b}
\end{table}

\begin{table}[htbp]
\small
\begin{tabular}{l|l|r|r|r|r|r}
\hline
\textbf{Study} & \textbf{Fluid (T\_bulk)} & \textbf{Wall Temp.} & \textbf{Nuc. Sites} & \textbf{T\_wall - T\_bulk} & \textbf{Sat. Temp.} & \textbf{Stefan Num.} \\
\hline
\hline
\multirow{20}{*}{Saturated} & \multirow{20}{*}{LN2 (-196\textdegree C)} & -191 & 5 & 5 & 0 & 0.0512 \\
 & & -190 & 6 & 6 & 0 & 0.0614 \\
 & & -188 & 8 & 8 & 0 & 0.0818 \\
 & & -186 & 10 & 10 & 0 & 0.1023 \\
 & & -184 & 12 & 12 & 0 & 0.1228 \\
 & & -182 & 14 & 14 & 0 & 0.1432 \\
 & & -180 & 16 & 16 & 0 & 0.1637 \\
 & & -178 & 18 & 18 & 0 & 0.1841 \\
 & & -176 & 20 & 20 & 0 & 0.2046 \\
 & & -175 & 21 & 21 & 0 & 0.2148 \\
 & & -174 & 22 & 22 & 0 & 0.2251 \\
 & & -172 & 24 & 24 & 0 & 0.2455 \\
 & & -170 & 26 & 26 & 0 & 0.2660 \\
 & & -168 & 28 & 28 & 0 & 0.2864 \\
 & & -166 & 30 & 30 & 0 & 0.3069 \\
 & & -165 & 31 & 31 & 0 & 0.3171 \\
 & & -164 & 32 & 32 & 0 & 0.3274 \\
 & & -162 & 34 & 34 & 0 & 0.3478 \\
 & & -160 & 36 & 36 & 0 & 0.3683 \\
 & & -159 & 37 & 37 & 0 & 0.3785 \\
\hline
\multirow{20}{*}{Subcooled} & \multirow{20}{*}{LN2 (-204\textdegree C)} & -181 & 3 & 23 & 0.3478 & 0.2353 \\
 & & -180 & 4 & 24 & 0.3333 & 0.2455 \\
 & & -178 & 6 & 26 & 0.3077 & 0.2660 \\
 & & -176 & 8 & 28 & 0.2857 & 0.2864 \\
 & & -174 & 10 & 30 & 0.2667 & 0.3069 \\
 & & -172 & 12 & 32 & 0.2500 & 0.3274 \\
 & & -170 & 14 & 34 & 0.2353 & 0.3478 \\
 & & -168 & 16 & 36 & 0.2222 & 0.3683 \\
 & & -166 & 18 & 38 & 0.2105 & 0.3887 \\
 & & -165 & 19 & 39 & 0.2051 & 0.3990 \\
 & & -164 & 20 & 40 & 0.2000 & 0.4092 \\
 & & -162 & 22 & 42 & 0.1905 & 0.4297 \\
 & & -160 & 24 & 44 & 0.1818 & 0.4501 \\
 & & -158 & 26 & 46 & 0.1739 & 0.4706 \\
 & & -156 & 28 & 48 & 0.1667 & 0.4910 \\
 & & -155 & 29 & 49 & 0.1633 & 0.5013 \\
 & & -154 & 30 & 50 & 0.1600 & 0.5115 \\
 & & -152 & 32 & 52 & 0.1538 & 0.5320 \\
 & & -150 & 34 & 54 & 0.1481 & 0.5524 \\
 & & -149 & 35 & 55 & 0.1455 & 0.5627 \\
\hline
\end{tabular}
\caption{Boiling Curve Data for LN2}
\label{tab:datasheet-ln2}
\end{table}

\end{document}